\documentclass[conference]{IEEEtran}
\usepackage{times}

\usepackage[numbers]{natbib}
\usepackage{multicol}
\usepackage[bookmarks=true]{hyperref}
\usepackage{amsmath, amsfonts, amssymb, amsthm}
\usepackage{enumerate}
\usepackage[inline]{enumitem}
\usepackage{mathtools}
\usepackage{graphicx}
\usepackage{longtable,tabularx}
\usepackage{placeins} 
\usepackage{float}
\usepackage{multirow}
\usepackage{bbm}
\usepackage[table, dvipsnames]{xcolor}
\usepackage{threeparttable}
\usepackage{balance}
\usepackage[ruled,algo2e]{algorithm2e}
\usepackage{algpseudocode}
\usepackage{adjustbox}
\usepackage{booktabs}
\usepackage{bm}
\usepackage{etoolbox}
\usepackage{microtype}
\usepackage{units}
\usepackage{cleveref}
\usepackage{caption}
\usepackage[textsize=tiny]{todonotes}

\newcommand{\diag}{\mathrm{\mathbf{diag}}}

\newcommand{\minimize}[1]{ \ensuremath{\underset{#1}{\mathrm{minimize}}\ }}

\newcommand*{\var}{\ensuremath{\mathrm{var}}}

\newcommand*{\trace}{\ensuremath{\mathrm{trace}}}

\newcommand{\mcal}[1]{\mathcal{#1}}
\newcommand{\mbb}[1]{\mathbb{#1}}
\newcommand{\T}{\mathsf{T}}

\newcommand{\norm}[1]{\left\Vert #1 \right\Vert}

\newcommand{\SE}{\ensuremath{\mathrm{SE}}}
\newcommand{\SO}{\ensuremath{\mathrm{SO}}}
\newcommand{\algname}{\mbox{SIREN}\xspace}

\newbool{extended}
\setbool{extended}{false}

\makeatletter
\newcommand{\oldnormaux}[3]{\mathpalette\oldnormaux@i{{#1}{#2}{#3}}}
\newcommand{\oldnormaux@i}[2]{\oldnormaux@ii#1#2}
\newcommand{\oldnormaux@ii}[4]{%
  \sbox\z@{$\m@th#1#2#4#3$}%
  \sbox\tw@{$\m@th\|$}%
  \mathopen{\hbox to\wd\tw@{\hss\vrule height \ht\z@ depth \dp\z@ width .2\wd\tw@\hss}}%
  #4
  \mathclose{\hbox to\wd\tw@{\hss\vrule height \ht\z@ depth \dp\z@ width .2\wd\tw@\hss}}%
}
\makeatother

\makeatletter
\newcommand{\longdash}[1][2em]{%
  \makebox[#1]{$\m@th\smash-\mkern-7mu\cleaders\hbox{$\mkern-2mu\smash-\mkern-2mu$}\hfill\mkern-7mu\smash-$}}
\makeatother
\newcommand{\omitskip}{\kern-\arraycolsep}

\newcommand{\insertfig}{%
\setcounter{figure}{0}
\begin{center}
    \includegraphics[width=\linewidth]{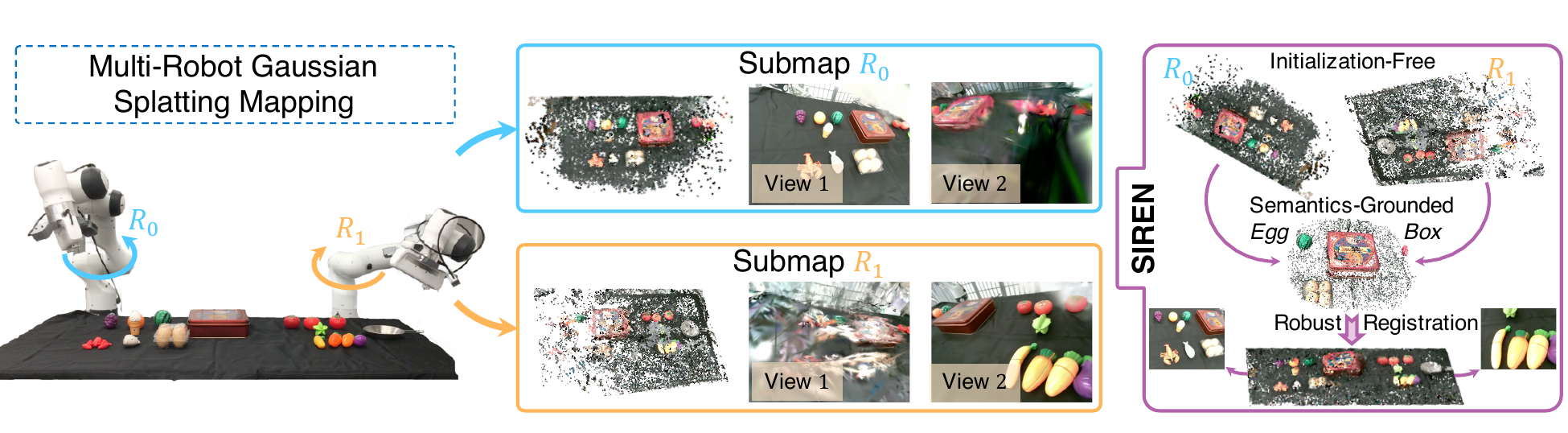}
    \captionof{figure}{\algname enables robust registration (i.e., fusion) %
    of multi-robot Gaussian Splatting maps, with no access to camera poses, images, and inter-map relative poses, via semantics-grounded optimization centered on feature-rich regions of each map.
    }%
    \label{fig:banner}
\end{center}
}

\makeatletter
\apptocmd{\@maketitle}{\centering\insertfig}{}{}%
\makeatother

\begin{document}

\title{\algname: Semantic, Initialization-Free Registration of Multi-Robot Gaussian Splatting Maps}

\author{Ola Shorinwa, Jiankai Sun, Mac Schwager, Anirudha Majumdar}

\maketitle

\begin{abstract}
  We present \algname for registration of multi-robot Gaussian Splatting (GSplat) maps, with \emph{zero} access to camera poses, images, and inter-map transforms for initialization or fusion of local submaps. To realize these capabilities, \algname harnesses the versatility and robustness of semantics in three critical ways to derive a rigorous registration pipeline for multi-robot GSplat maps. First, \algname utilizes semantics to identify feature-rich regions of the local maps where the registration problem is better posed, eliminating the need for any initialization which is generally required in prior work. Second, \algname identifies candidate correspondences between Gaussians in the local maps using robust semantic features, constituting the foundation for robust geometric optimization, coarsely aligning 3D Gaussian primitives extracted from the local maps. Third, this key step enables subsequent photometric refinement of the transformation between the submaps, where \algname leverages novel-view synthesis in GSplat maps along with a semantics-based image filter to compute a high-accuracy non-rigid transformation for the generation of a high-fidelity fused map. We demonstrate the superior performance of \algname compared to competing baselines across a range of real-world datasets, and in particular, across the most widely-used robot hardware platforms, including a manipulator, drone, and quadruped. In our experiments, \algname achieves about $90$x smaller rotation errors, $300$x smaller translation errors, and $44$x smaller scale errors in the most challenging scenes, where competing methods struggle.
  We will release the code and provide a link to the project page after the review process.
\end{abstract}


\section{Introduction}
\label{sec:introduction}
In robotics, traditional map representations such as point-cloud and voxel maps constitute a critical component of the robotics stack, enabling downstream behavior prediction, planning, and control across many problem domains, e.g., navigation and manipulation. However, these map representations often lack the expressiveness required to capture high-fidelity visual details and semantics \cite{kerbl20233d}, limiting their applications in fine-grained robotics tasks, e.g., in dexterous open-vocabulary manipulation \cite{shen2023distilled}. To address these fundamental limitations, recent robotics research has adopted radiance fields as flexible, high-fidelity $3$D scene representations, e.g., in robot navigation \cite{chen2024splat, qiu2024learning} and manipulation \cite{rashid2023language, shorinwa2024splat}. Radiance fields, e.g., neural radiance fields (NeRFs) \cite{mildenhall2021nerf} and Gaussian Splatting (GSplat) \cite{kerbl20233d}, are trained entirely from monocular images, typically collected by a single robot on a single deployment.

However, practical real-world robot mapping requires multiple deployments and multiple robot platforms, especially when mapping large-scale areas. For example, mobile robots have a limited battery life, while fixed-base robotic manipulators have a limited workspace, making map registration a necessity for covering large-scale areas.
Fusing map information across multiple robot platforms and deployments remains a key challenge, particularly with radiance field maps.
Prior work has explored fusing multiple radiance field maps \cite{chang2025gaussreg, yuan2024photoreg}; however, these methods either require a good initialization of inter-map correspondences or access to the camera poses and images, which is often unavailable in many practical situations.
Moreover, these methods often fail in unstructured real-world environments, an important operational domain for robots.
To address these challenges, we introduce \emph{\algname}, a semantic, initialization-free registration algorithm for multi-robot Gaussian Splatting maps.

Although often unexploited, many real-world scenes contain rich semantic information, e.g., associated with objects such as vehicles, people, utensils, and vegetation. Leveraging this key insight, \algname trains a semantic GSplat to directly embed semantic features in GSplat maps and subsequently uses the inherent semantics in the local maps to identify feature-rich regions of the local maps, providing a more reliable set of Gaussians for the identification of candidate correspondences. This critical design choice underpins the superior performance of \algname. Specifically, the core challenge in map registration can be largely attributed to the difficulty in identifying accurate correspondences between points \cite{tam2012registration}. In fact, given accurate correspondences, the map registration problem can be solved efficiently in closed-form. By centering the registration problem on feature-rich regions of the local maps to derive a reliable set of correspondences, \algname addresses this core challenge. Subsequently, \algname harnesses the robustness of semantic features to formulate a geometric optimization problem for a coarse non-rigid relative transformation aligning Gaussian primitives across the local maps. We solve the geometric optimization problem efficiently in closed-form. Although the coarsely aligned map may be satisfactory in certain scenarios, the fused map often lacks the photorealism afforded by GSplat  maps. To address this weakness, \algname leverages novel-view synthesis of GSplat maps to render images across the local maps, used as a supervision signal for computing a high-accuracy relative transformation. To guard against the impacts of inaccurate renderings from the local GSplats, we utilize a semantics-based image filter to identify reliable candidate images, which we use for supervision.
Consequently, \algname generates photorealistic fused GSplat maps from the local multi-robot maps, illustrated in \Cref{fig:banner}, where we show the key components of \algname.

We demonstrate the superior effectiveness of \algname compared to both existing GSplat registration methods and classical point-cloud registration methods across different real-world datasets, including standard benchmarks for radiance fields and data collected across three different robot hardware platforms: a quadruped, drone, and fixed-base manipulator. In almost all settings, \algname achieves lower rotation, translation, and scale errors compared to all baselines, especially in the quadruped mapping task, where \algname achieves about $90$x lower rotation error, $300$x lower translation error, and $44$x lower scale errors. We summarize our contributions:
\begin{itemize}
    \item We introduce a \emph{semantics-grounded} feature extraction and matching method for GSplat map registration, centering the registration problem on feature-rich regions, addressing a key challenge of registration algorithms.
    \item We derive a Gaussian-to-Gaussian registration procedure for coarse alignment, utilizing \emph{semantic correspondence} to identify and mitigate the effects of outliers in the registration process.
    \item We present a photometric registration procedure, leveraging novel-view synthesis of GSplats and a \emph{semantics-based} image filter to compute high-accuracy relative transformations, generating photorealistic fused maps.
    \item Together, these components constitute \algname, enabling the registration of multi-robot GSplat maps, with \emph{zero} access to source images or poses and no inter-map relative pose initialization.
\end{itemize}

\section{Related Work}
\label{sec:related_work}

\smallskip
\noindent\textbf{Radiance Fields.}
Neural radiance fields (NeRFs) \cite{mildenhall2021nerf} significantly outperform traditional $3$D scene reconstruction methods, such as those based on point clouds and voxels, generating photorealistic renderings, which capture intricate levels of geometric and visual details. NeRFs represent a scene using volumetric density and color fields over a $5$D input space, comprising a $3$D location and a $2$D viewing direction. NeRFs parameterize each field using multi-layer perceptrons (MLPs) trained through gradient descent. Although NeRFs achieve remarkable high-fidelity reconstructions, NeRFs are limited by significant training time and slow rendering speeds \cite{zhang2020nerf++, yu2021pixelnerf, barron2022mip}. Gaussian Splatting \cite{kerbl20233d} was introduced to address these limitations. GSplats represent the scene using ellipsoidal primitives, each with a mean and covariance (spatial and geometric parameters) and opacity and spherical harmonic parameters (visual-related parameters). GSplats generate high-fidelity scene renderings at real-time speeds with generally faster training times compared to NeRFs. Recent work has improved the geometric accuracy of GSplats \cite{guedon2024sugar, huang20242d}, in addition to eliminating high-frequency artifacts \cite{yu2024mip, lee2025deblurring}.

\smallskip
\noindent\textbf{Semantic Radiance Fields.}
Large vision-language models, e.g., CLIP \cite{radford2021learning} and DINO \cite{caron2021emerging, oquab2023dinov2} have demonstrated the effectiveness of large-scale pretraining in learning robust visual and language features, enabling object detection \cite{gu2021open, minderer2022simple}, object segmentation \cite{wang2022cris, luddecke2022image}, and image captioning \cite{mokady2021clipcap, luo2022clip4clip}. Prior work has examined grounding the $2$D image-language features from vision-language foundation models in $3$D radiance fields. CLIP-NeRF \cite{wang2022clip}, DFF \cite{kobayashi2022decomposing}, and LERF \cite{kerr2023lerf} train NeRFs with CLIP image-language features, enabling open-vocabulary object segmentation and scene-editing. Similarly, subsequent work has enabled distillation of semantic features into GSplats \cite{qin2024langsplat, zhou2024feature}, with similar open-vocabulary object segmentation quality, albeit at much faster rendering rates \cite{shorinwa2024fast}. Moreover, prior work has leveraged semantic radiance fields to enable GSplat-based world models \cite{lu2025manigaussian} and open-vocabulary robotic manipulation in NeRFs \cite{shen2023distilled, rashid2023language} and GSplat environments \cite{shorinwa2024splat, ji2024graspsplats}. In this work, we leverage semantic radiance fields for registration of $3$D maps, which has not been explored in prior work, to the best of our knowledge.

\smallskip
\noindent\textbf{Point Cloud Registration.}
The Iterative Closest Point (ICP) algorithm \cite{besl1992method} has proven to be notably effective for point cloud registration, despite its simplicity. However, ICP generally requires a good initial solution, which is often computed using global registration techniques, e.g., RANSAC \cite{fischler1981random, holz2015registration} and FGR \cite{zhou2016fast}. Many variants of ICP have been introduced to improve its robustness \cite{chen1992object, rusinkiewicz2001efficient, bouaziz2013sparse, park2017colored}, leveraging the local color and geometry of the constituent points for faster convergence. More recently, learning-based methods \cite{wang2019deep, fu2021robust, qin2023geotransformer} have emerged for point cloud registration, utilizing convolutional neural networks (CNNs) and transformers for feature extraction and feature matching to compute the correspondences between points.

\smallskip
\noindent\textbf{Registration of Radiance Fields.}
Training large-scale radiance fields is often infeasible, due to computational resource constraints. 
Consequently, Nerf2nerf \cite{goli2023nerf2nerf} aligns individually-trained NeRFs with different frames into a shared reference frame, by extracting the geometry of the scene from the NeRF as a surface field. Nerf2nerf requires human annotation of keypoints within each NeRF for registration, posing a practical challenge.
Similarly, the NeRF registration methods in \cite{jiang2023registering, chen2023dreg} computes spatial features of NeRFs using learned feature descriptors $3$D primitives and subsequently estimates the transformation between the source and target NeRFs using the Kabsch-Umeyama algorithm \cite{umeyama1991least} or RANSAC.

More recent work has explored the registration of GSplats. 
LoopSplat \cite{zhu2024loopsplat} and PhotoReg \cite{yuan2024photoreg} compute the optimal transformation between GSplat maps by minimizing the rendering loss but require access to the set of camera poses (keyframes) of each GSplat. In contrast, GaussReg \cite{chang2025gaussreg} computes a coarse transformation between two GS-maps using a geometric transformer \cite{qin2023geotransformer}, which is refined with a 2D convolutional neural network augmented with a geometric transformer, without access to the camera poses. In contrast to these existing methods, \algname leverages the semantics inherent to GSplat maps to identify regions of overlap and to coarsely align GSplat maps, eliminating the need for access to camera poses or images. Moreover, unlike GaussReg, \algname does not require a separate training procedure for the learned CNN and geometric transformer models.

\section{Preliminaries}
\label{sec:preliminaries}
We present relevant notation used in this paper. We denote the trace of a function by $\trace$ and the determinant and variance of a matrix by $\det$ and $\var$, respectively. Further, we denote the strictly positive orthant by ${\mbb{R}_{++}}$. Next, we provide a brief introduction to Gaussian Splatting.
Gaussian Splatting represents non-empty space in a scene using a set of ellipsoidal primitives, each parameterized by a mean ${\mu \in \mbb{R}^{3}}$, a covariance ${\Sigma \in \mbb{R}^{3 \times 3}}$ defined by a rotation matrix ${H \in \SO(3)}$ and a diagonal scaling matrix ${\Lambda \in \mbb{R}^{3 \times 3}}$, an opacity parameter ${\alpha \in [0, 1]}$, and spherical harmonic parameters. These attributes are optimized via gradient descent on the loss function: ${\mcal{L}_{\mathrm{gs}} = (1 - \lambda) \sum_{\mcal{I} \in \mcal{D}} \norm{\mcal{I} - \hat{\mcal{I}}}_{1} + \lambda \mcal{L}_{\mathrm{D-SSIM}}}$, over the training dataset $\mcal{D}$, where ${\lambda \in (0, 1)}$ represents the relative weight term and $\mcal{L}_{\mathrm{D-SSIM}}$ represents the differentiable structural similarity loss index measure. The first term in the rendering loss represents the photometric loss between the ground-truth image and the rendered image, generated via a tile-based rasterization procedure, given a camera pose.

\section{\algname for GSplat Map Registration}
\label{sec:method}
\begin{figure*}[th]
    \centering
    \includegraphics[width=\linewidth]{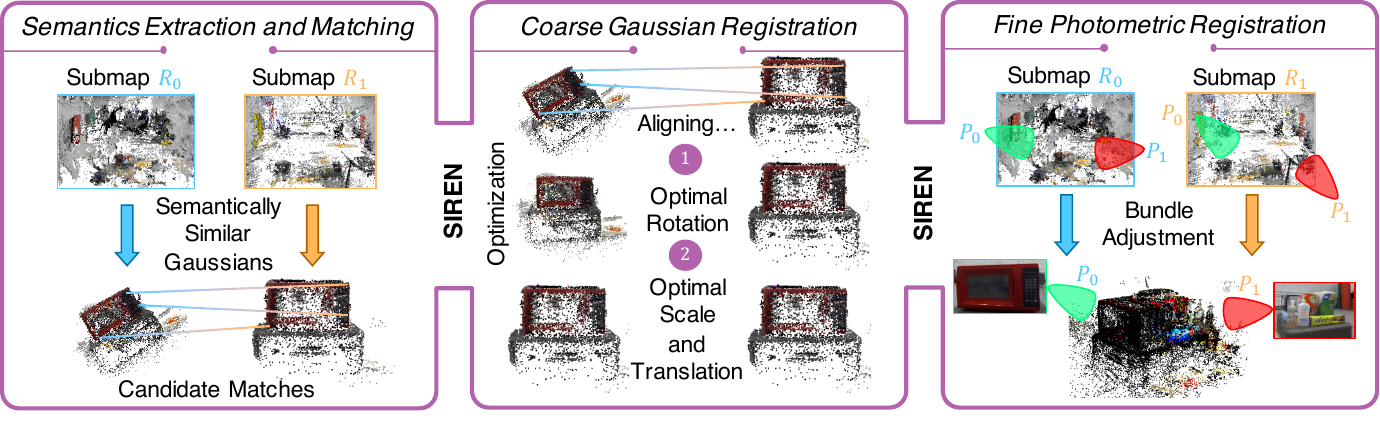}
    \caption{\algname consists of three steps: (a) semantic feature extraction and matching of Gaussians across the local maps, (b) coarse Gaussian-to-Gaussian registration for coarsely aligning the local maps, (c) fine photometric registration for high-accuracy fusion of the local maps, through image-to-image registration and bundle adjustment.}
    \label{fig:siren_pipeline}
\end{figure*}

We introduce \algname, our semantics-grounded, initialization-free registration algorithm for GSplat maps. At its core, \algname leverages open-vocabulary semantics within a principled optimization-based framework to enable the robust registration of multi-robot GSplat maps. \algname utilizes an SCF procedure, composed of: (a) Semantic feature extraction and matching, (b) Coarse Gaussian-to-Gaussian geometric registration, and (c) Fine photometric registration, illustrated in 
\Cref{fig:siren_pipeline}.
Here, for simplicity, we discuss the registration pipeline in the problem setting with two multi-robot maps, where we seek to register a source GSplat map to a target GSplat map. However, the discussion applies to the registration of multiple local multi-robot maps.

In the first step, \algname identifies corresponding pairs of Gaussians in a pair of GSplat maps by examining the similarity between the semantic features of the ellipsoids. Subsequently, given the set of corresponding Gaussians, \algname solves a Gaussian-to-Gaussian optimization problem to compute the optimal transformation aligning the pair of multi-robot maps with a robust objective function, which leverages the semantic similarity between each pair of ellipsoids to guard against the impacts of outliers. In the last step, \algname harnesses the novel-view synthesis capabilities of Gaussian Splatting to render candidate images for image-to-image registration, enabling fine registration of both maps via a structure-from-motion-based approach. In this stage, \algname utilizes image-level semantic features to identify pairs of corresponding images, critical to the robust matching of local features such as corners, edges, and blobs between the images. We discuss each of these steps in greater detail.

\subsection{Semantic Feature Extraction and Matching}
Noting that semantics underpin \algname, we begin with a discussion of the semantic distillation procedure utilized by \algname in grounding $2$D semantic information from the vision-language model in GSplat maps, where we associate semantic embeddings with each ellipsoid in the GSplat map. 

\smallskip
\noindent\textbf{Semantic Gaussian Splatting.}
Existing methods for training semantic GSplat models generally train auxiliary models, e.g., autoencoders or CNNs, for dimensionality reduction of the semantic features from vision-language models to compute lower-dimensional semantic features, which are distilled into the GSplat model \cite{qin2024langsplat, zhou2024feature}. These methods require relatively significant computation time and GPU memory, which we seek to avoid in \algname. Consequently, we take a different approach to semantic distillation. In \algname, we simultaneously train a semantic field $\psi$ alongside the GSplat model. The semantic field ${\psi: \mbb{R}^{3} \mapsto \mbb{R}^{d}}$ maps $3$D points to $d$-dimensional semantic features, where $d$ is determined by the vision-language foundation model, e.g., ${d = 512}$~or~${1024}$ in CLIP. We parameterize ${\psi}$ with a multi-resolution neural hashgrid, trained with the GSplat model, using the loss function:
\begin{equation}
    \label{eq:semantics_loss_function}
    \mcal{L} = \mcal{L}_{\mathrm{gs}} + \gamma \sum_{\mcal{I} \in \mcal{D}} \norm{\mcal{I}_{f} - \hat{\mcal{I}}_{f}}_{F}^{2} - \beta \sum_{\mcal{I} \in \mcal{D}} \phi(\mcal{I}_{f}, \hat{\mcal{I}}_{f}),
\end{equation}
where ${\mcal{I}_{f} \in \mbb{R}^{W \times H \times d}}$ and ${\hat{\mcal{I}}_{f} \in \mbb{R}^{W \times H \times d}}$ represent the ground-truth and predicted semantic feature maps associated with each image in the training dataset $\mcal{D}$, $\gamma$ and $\beta$ represent relative weight terms, and $\phi$ represents the cosine-similarity function between each semantic feature in $\mcal{I}_{f}$ and $\hat{\mcal{I}}_{f}$. 

To predict the semantic feature map associated with each training image, we leverage a key insight of Gaussian Splatting: Gaussian Splatting provides highly-accurate depth estimates, even without any depth supervision \cite{huang20242d}. This key insight enables us to avoid training proposal networks (as required in NeRFs) that generate samples of the termination points of rays associated with each pixel in the rendered image of a camera, ultimately enabling \algname to avoid significant compute and memory overhead associated with training proposal networks.
As such, given a camera pose, we back-project points from the image plane to the $3$D world and pass these points into $\psi$ to predict the semantic feature associated with these points. We augment each pixel in the image plane with its semantic features to obtain the semantic feature map. Training the GSplat with the semantic component does not adversely impact the photometric performance of the GSplat, enabling us to utilize the same hyperparameters and adaptive densification procedure used in the original GSplat work \cite{kerbl20233d}.

\smallskip
\noindent\textbf{Feature Extraction.}
In the feature extraction and matching step, \algname identifies feature-rich areas of the scene via semantic localization, to improve the robustness of the subsequent optimization-based registration steps, as the feasibility and convergence of the optimization problems significantly depend on the presence of informative features.
Given a trained semantic GSplat model, we augment each Gaussian with a semantic attribute, computed by querying the semantic field $\psi$ at the mean $\mu$ of each Gaussian. Subsequently, from a set of open-vocabulary queries, we compute the semantic relevancy score between each Gaussian and the natural-language query by taking the pairwise softmax over the cosine-similarity between the semantic feature of each Gaussian and the semantic embedding associated with the text query and the cosine-similarity between the semantic feature of each Gaussian and the semantic embedding associated with a generic or null text query (i.e., a text query for a generic object or an object a user does not want to localize) \cite{kerr2023lerf}. Depending on the quality of the local multi-robot maps, we can post-process the resulting set of Gaussians to either inflate the set by incorporating other Gaussians in close proximity to the initial set (based on the geometric or semantic distance) or deflate the set by removing Gaussians considered to be outliers (based on statistics, e.g., the standard deviation of the distance between neighboring Gaussians).

\smallskip
\noindent\textbf{Feature Matching.}
Given the set of extracted Gaussians for each map, we match Gaussians from the source GSplat map to the target GSplat map, resulting in a set of correspondences $\mcal{E}$ where ${(i, j) \in \mcal{E}}$ indicates that Gaussian $i$ in the source GSplat map corresponds to Gaussian $j$ in target GSplat map. To identify the candidate matches in $\mcal{E}$, we compute the cosine-similarity between the semantic embeddings of the Gaussians in the source and target maps. We match each Gaussian in the source map to a random set of $M$ Gaussians in the target map, sampling among the target Gaussians (i.e., Gaussians in the target map) that are within a specified distance from the source Gaussian (i.e., Gaussian in the source map). In the sampling step, \algname can utilize a uniform distribution or a distribution where the probability values are proportional to the cosine-similarity values between the source and target Gaussians. For computational reasons, this operation can be performed using efficient data structures such as KD-trees. In addition, we can repeat the procedure to match each Gaussian in the target GSplat map to a set of Gaussians in the source GSplat map, taking care to ensure that $\mcal{E}$ does not contain any duplicate entry. Moreover, the matching process can be augmented with geometric information, by selecting candidate matches using geometric descriptors, such as the Fast Point Feature Histograms (FPFH) descriptors \cite{rusu2009fast}. We denote the set of Gaussians in the source map present in $\mcal{E}$ by $\mcal{P}$ and the set of Gaussians in the target map present in $\mcal{E}$ by $\mcal{Q}$.

\subsection{Coarse Gaussian-to-Gaussian Registration}
\label{sec:coarse_gaussian_to_gaussian_registration}
We use the output from the feature matching step to compute an initial non-rigid transformation, consisting of a scale ${s_{c} \in \mbb{R}}$, a rotation matrix ${R \in \SO(3)}$, and a translation vector ${t \in \mbb{R}^{3}}$, aligning the Gaussians in the source GSplat map to the Gaussians in the target GSplat map. Since computing this transformation using all the Gaussians is intractable in general, we solve for this transformation using the Gaussians in $\mcal{P}$ and $\mcal{Q}$, a feature-dense, much smaller set of Gaussians, a design choice that not only reduces the computational cost, but also improves the feasibility and convergence properties of the resulting optimization problem. Specifically, we formulate the coarse Gaussian-to-Gaussian registration problem as an optimization problem over the transformation parameters, given by:
\begin{equation}
    \label{eq:coarse_registration_nonsimplified}
    \begin{aligned}
        \minimize{s_{c} \in \mbb{R}_{++}, R \in \SO(3), t \in \mbb{R}^{3}} \frac{1}{2} \sum_{(i, j) \in \mcal{E}} & w_{ij} \left(\norm{s_{c}Rp_{i} + t - q_{j}}_{2}^{2} \right. \\
        &\quad \left. + \norm{s_{c}^{2}R \Sigma_{p_{i}}R^{\T} - \Sigma_{q_{j}}}_{F}^{2} \right),
    \end{aligned}
\end{equation}
where ${p_{i} \in \mbb{R}^{3}}$ denotes the mean of Gaussian $i$ in the source map,  ${q_{j} \in \mbb{R}^{3}}$ denotes the mean of Gaussian $j$ in the target map, and $\Sigma_{p_{i}}$ and $\Sigma_{q_{j}}$ denote the covariance of Gaussian $i$ in the source map and Gaussian $j$ in the target map, respectively. We introduce the weight $w_{ij}$ in \eqref{eq:coarse_registration_nonsimplified} to increase the robustness of the optimization problem to outliers (i.e., false correspondences), by reducing the influence of outliers. We define $w_{ij}$ to be proportional to the cosine-similarity between the semantic embeddings of Gaussian $i$ in $\mcal{P}$ and Gaussian $j$ in $\mcal{Q}$. The optimization problem in \eqref{eq:coarse_registration_nonsimplified} is challenging to solve in general, necessitating the derivation of a less challenging formulation. Noting that the covariance of the Gaussians in the source and target maps can be expressed in the form ${\Sigma_{p_{i}} = H_{p_{i}} \Lambda_{p_{i}} \Lambda_{p_{i}}^{\T} H_{p_{i}}^{\T}}$ and ${\Sigma_{q_{j}} = H_{q_{j}} \Lambda_{q_{j}} \Lambda_{q_{j}}^{\T} H_{q_{j}}^{\T}}$, where $H_{p_{i}}$ and $H_{q_{j}}$ denote the orientation of Gaussian $i$ in the source map and Gaussian $j$ in the target map, respectively, and $\Lambda_{p_{i}}$ and $\Lambda_{q_{j}}$ denote the scale of the Gaussians, we express the problem in \eqref{eq:coarse_registration_nonsimplified} in the form:
\begin{equation}
    \label{eq:coarse_registration}
    \begin{aligned}
        \minimize{s_{c} \in \mbb{R}_{++}, R \in \SO(3), t \in \mbb{R}^{3}} \frac{1}{2} \sum_{(i, j) \in \mcal{E}} & w_{ij} \left(\norm{s_{c}Rp_{i} + t - q_{j}}_{2}^{2} \right. \\
        & \left. + \norm{s_{c}R H_{p_{i}}  \Lambda_{p_{i}} - H_{q_{j}}  \Lambda_{q_{j}}}_{F}^{2} \right),
    \end{aligned}
\end{equation}
which can be solved efficiently in closed-form, which we show in Appendix~\ref{app:method}, with:
\begin{align}
    R_{c}^{\star} &= U_{c} \Theta_{c} V_{c}^{\top}, \label{eq:coarse_registration_opt_R} \\
    s_{c}^{\star} &= \frac{\trace(\Theta_{c} \Sigma)}{\trace\left(W\check{P}^{\top}\check{P} + \sum_{(i, j) \in \mcal{E}} w_{ij} \check{H}_{p_{i}}^{\top}\check{H}_{p_{i}}\right)}, \label{eq:coarse_registration_opt_scale} \\
    t_{c}^{\star} &= \tilde{\mu}_{\mcal{Q}} - s_{c}^{\star} R^{\star} \tilde{\mu}_{\mcal{P}}, \label{eq:coarse_registration_opt_translation}
\end{align}
where ${U_{c} \Sigma_{c} V_{c}^{\top} = \check{Q}W\check{P}^{\top}
+ \sum_{(i, j) \in \mcal{E}} w_{ij} \check{H}_{q_{j}}\check{H}_{p_{i}}^{\top}}$, computed via the singular value decomposition (SVD),
and ${\Theta_{c} = \diag(1, 1, \det(U_{c}V_{c}^{\top}))}$.
We define ${\tilde{\mu}_{\mcal{P}}}$ and ${\tilde{\mu}_{\mcal{Q}}}$ as the weighted average of the means of the Gaussians in $\mcal{P}$ and $\mcal{Q}$, with weights $w_{ij}$ for Gaussian $i$ in $\mcal{P}$ and Gaussian $j$ in $\mcal{Q}$. Further, ${\check{P} \in \mbb{R}^{3 \times N}}$ and ${\check{Q} \in \mbb{R}^{3 \times N}}$ represent the \emph{zero-centered} Gaussians in $\mcal{P}$ and ${\mcal{Q}}$, respectively, with the $i$th column of $\check{P}$ given by ${\check{P}_{i} = p_{i} - \tilde{\mu}_{\mcal{P}}}$ and similarly for the $j$th column of $\check{Q}$.  We introduce the terms ${\check{H}_{p_{i}} \in \mbb{R}^{3 \times 3}}$ and ${\check{H}_{q_{j}} \in \mbb{R}^{3 \times 3}}$ to simplify notation, with: ${\check{H}_{p_{i}} = H_{p_{i}}  \Lambda_{p_{i}}}$ and ${\check{H}_{q_{j}} = H_{q_{j}}  \Lambda_{q_{j}}}$. In addition, ${W \in \mbb{R}^{N \times N}}$ denotes the diagonal weight matrix, ${W_{kk} = w_{k}}$, with ${w_{k} = w_{ij}}$,~${\forall k = (i, j) \in \mcal{E}}$.

Although the resulting solution is optimal for the problem in \eqref{eq:coarse_registration}, the solution of \eqref{eq:coarse_registration} might not be optimal for the registration of the two sets of Gaussians, i.e., $\mcal{P}$ and $\mcal{Q}$, given that $\mcal{C}$ might contain spurious correspondences. To improve the robustness of \algname to spurious correspondences, we utilize RANSAC \cite{fischler1981random} when solving the optimization problem in \eqref{eq:coarse_registration}. With RANSAC, we iteratively update the correspondences in $\mcal{C}$ to remove false correspondences and compute an optimal transformation associated with the resulting set of correspondences.

\subsection{Fine Photometric Registration}
In the preceding coarse registration step, \algname computes a transformation aligning the source and target maps using only the geometric attributes of the Gaussians in each map. The coarse registration step fails to leverage the highly-informative visual features inherent in the GSplat maps, effectively limiting the accuracy of the estimated transformation. To overcome this limitation, \algname harnesses the novel-view synthesis capability of GSplat maps to generate photorealistic images and optimizes over the resulting set of rendered images to compute a transformation consistent with the rendered images from the source and target maps. The fine photometric registration procedure employs a lightweight structure-from-motion framework to minimize the computation costs, while improving the fidelity of the registered maps. This procedure consists of the following steps: (i) image generation, (ii) image registration and triangulation, and (iii) bundle adjustment, which we discuss in the rest of this section.

\smallskip
\noindent\textbf{Image Generation and Matching.}
The fine registration procedure begins with the identification of a set of images with common features across the source and target maps, constituting arguably the most important step of the fine registration procedure. In particular, the feasibility of the fine registration procedure hinges on matching corresponding features across all images in the set. In general, identifying good candidate images for the matching process is challenging, especially without any prior knowledge of the region of overlap between the source and target maps. To address this challenge, we leverage the semantic submap extracted in the first stage of \algname to identify a region of overlap between the source and target maps. Subsequently, we exploit novel-view synthesis in Gaussian Splatting to render images at corresponding poses in both maps, by transforming the camera pose in one map to the associated camera pose in the other map, utilizing the coarse registration result to compute the corresponding pose. With this approach, not only do the resulting images contain common features from the overlapping region, the images also contain a dense set of features, associated with the semantic submap. However, the pair of rendered images may not contain sufficient matches, which could degrade the accuracy of the fine registration procedure. To mitigate this risk, we harness image semantics in vision foundation models to evaluate the similarity between each pair of rendered images, retaining only sufficiently similar images. In this work, we use CLIP along with the cosine-similarity metric, given that the image embeddings of CLIP were trained with a cosine-similarity loss function; however, other vision foundation can also be used, e.g., \cite{caron2021emerging}.

\smallskip
\noindent\textbf{Image Registration and Triangulation.}
Following the generation of corresponding images, we extract features from all images using the learned feature extractors NetVLad \cite{arandjelovic2016netvlad} for global image-level descriptors and SuperPoint \cite{sarlin2020superglue} for local features, which we found to be more robust compared to classical feature extractors, e.g., SIFT \cite{karami2017image}. Subsequently, we match features across all images using \cite{sarlin2020superglue}. From corresponding features, we estimate the relative pose of the camera and the estimated $3$D locations of the feature points via image registration and triangulation, yielding an initial estimate of the camera pose associated with each image in a common reference frame.

\smallskip
\noindent\textbf{Bundle Adjustment.}
The image registration step does not always provide high-accuracy camera pose estimates. Hence, we refine the estimated camera poses via bundle adjustment, i.e., we optimize over the camera pose and the $3$D locations of the feature points jointly through non-linear optimization. For brevity, we do not discuss the bundle adjustment problem in greater detail, noting its extensive discussion in prior work, e.g., \cite{schonberger2016structure}.  Although non-convex, the optimization problem can be solved efficiently via iterative methods, such as the Levenberg-Marquardth method, which we employ in this work. From the bundle adjustment optimization problem, we compute the camera poses associated with each image in an arbitrary common frame $\mcal{B}$. Given the camera poses expressed in $\mcal{A}$ and the corresponding poses in the source and target maps, we can compute an optimal transformation for registering $\mcal{A}$ to either the source frame (frame $\mcal{B}_{s}$) or the target frame (frame $\mcal{B}_{t}$) from the following registration problem in $\SE(3)$:
\begin{equation}
    \label{eq:fine_registration_nonsimplified}
    \begin{aligned}
        \minimize{s_{f} \in \mbb{R}_{++}, R \in \SO(3), t \in \mbb{R}^{3}} \frac{1}{2} \sum_{(i, j) \in \mcal{V}} & \left(\norm{s_{f}Ra_{i} + t - b_{j}}_{2}^{2} \right. \\
        &\quad \left. + \beta_{ij} \norm{R R_{c_{i}} - R_{d_{j}}}_{F}^{2} \right),
    \end{aligned}
\end{equation}
where ${s_{f}}$, ${R}$, and ${t}$ denote the scale, rotation, and translation parameters, respectively, $\mcal{V}$ denotes the set of edges between the camera poses expressed in $\mcal{A}$ and the corresponding poses in either the source or target frame, with ${a_{i} \in \mbb{R}^{3}}$ and ${b_{j} \in \mbb{R}^{3}}$ denoting the origin of the camera in $\mcal{A}$ and the origin of the camera in the frame $\mcal{B}_{s}$ or $\mcal{B}_{t}$, respectively, and ${R_{a_{i}}}$ and ${R_{b_{j}}}$ denoting the associated rotation matrices. We introduce the weight parameter ${\beta_{ij} \in \mbb{R}_{++}}$, which determines the contribution of the rotation-error component. In general, the optimization problem in \eqref{eq:fine_registration_nonsimplified} cannot be solved in closed-form. Solving \eqref{eq:fine_registration_nonsimplified} generally requires an iterative optimization method, e.g., sequential convex programming methods or Riemannian optimization methods.
However, as ${\beta_{ij}}$ approaches zero,~${\forall (i, j) \in \mcal{V}}$, the optimal solution \eqref{eq:fine_registration_nonsimplified} approaches a limit point, with:
\begin{equation}
    \label{eq:bundle_adjustment_opt}
    \begin{aligned}
        &R_{f}^{\star} \rightarrow U_{f} \Theta_{f} V_{f}^{\top}, 
        \enspace 
        s_{f}^{\star} \rightarrow \frac{\trace(\Theta_{f} \Sigma_{f})}{\trace\left(\check{A}^{\top}\check{A}\right)},
        \\
        &t_{f}^{\star} \rightarrow  \mu_{\mcal{B}} - s_{f}^{\star} R_{f}^{\star} \mu_{\mcal{A}},
    \end{aligned}
\end{equation}
where ${U_{f} \Sigma_{f} V_{f}^{\top} = \check{B}\check{A}^{\top}}$, ${\Theta_{f} = \diag(1, 1, \det(U_{f}V_{f}^{\top}))}$, $\mu_{\mcal{A}}$ and $\mu_{\mcal{B}}$ denote the mean of the camera origins in frames $\mcal{A}$ and $\mcal{B}$, respectively, and the $i$th column of ${\check{A} \in \mbb{R}^{3 \times N}}$ and the $j$th column of ${\check{B} \in \mbb{R}^{3 \times N}}$ are given by ${a_{i} - \mu_{\mcal{A}}}$ and ${b_{j} - \mu_{\mcal{B}}}$, respectively. The limit point follows from the derivation in \Cref{sec:coarse_gaussian_to_gaussian_registration} and \cite{umeyama1991least}. We can compose the pairwise transformations between frame $\mcal{A}$ and frames $\mcal{B}_{s}$ and $\mcal{B}_{t}$ to compute a transformation from $\mcal{B}_{s}$ and $\mcal{B}_{t}$. We apply the resulting transformation to the source map to express the source and target maps in a common frame and subsequently merge the resulting maps to obtain a composite GSplat map. Following the registration procedures, the composite map can be finetuned with new or existing data, which we explore in our experiments in Appendix~\ref{ssec:finetuning}.
We summarize the procedures in \algname in \Cref{alg:coarse_gaussian_registration}.

\begin{algorithm2e}[th]
    \caption{\algname: Multi-Robot Map Registration}
    \label{alg:coarse_gaussian_registration}
    
    \KwIn{Local GSplat Maps $\mcal{G}_{1}$, $\mcal{G}_{2}$\;}
    \KwOut{Fused GSplat Map $\mcal{G}_{f}$\;}
    \SetCommentSty{algsupercommfont}
    \tcp{Semantic Feature Extraction and Matching}
    Correspondence Set $\mcal{C} \gets \mathrm{GetCorrespondence}(\mcal{G}_{1}, \mcal{G}_{2})$\;    
    \tcp{Coarse Registration}
    \SetCommentSty{algcommfont}
    \tcp{Compute the Optimal Rotation}
    ${R_{c}^{\star} \gets \mathrm{Procedure~} \eqref{eq:coarse_registration_opt_R}}$\;
    \tcp{Compute the Optimal Scale}
    ${s_{c}^{\star} \gets \mathrm{Procedure~} \eqref{eq:coarse_registration_opt_scale}}$\;
    \tcp{Compute the Optimal Translation}
    ${t_{c}^{\star} \gets \mathrm{Procedure~} \eqref{eq:coarse_registration_opt_translation}}$\;
    \SetCommentSty{algsupercommfont}
    \tcp{Fine Registration}
    \SetCommentSty{algcommfont}
    \tcp{Get Images}
    ${\mcal{D}_{s} \gets \mathrm{Render} (\mcal{G}_{1}, \mcal{G}_{2}, R_{c}^{\star}, s_{c}^{\star}, t_{c}^{\star})}$\;
    \tcp{Refine Transformation}
    ${(R_{f}^{\star}, s_{f}^{\star}, t_{f}^{\star}) \gets \mathrm{Procedure~} \eqref{eq:bundle_adjustment_opt}}$\;
    \SetCommentSty{algsupercommfont}
    \tcp{Fuse Local Maps}
    \SetCommentSty{algcommfont}
    ${\mcal{G}_{f} \gets \mathrm{Fuse}(\mcal{G}_{1}, \mcal{G}_{2}, R_{f}^{\star}, s_{f}^{\star}, t_{f}^{\star})}$\;
\end{algorithm2e}

\section{Experiments}
\label{sec:evaluation}
We examine the performance of \algname in comparison to existing registration methods for Gaussian Splatting and point clouds. Specifically, we compare two variants of \algname---i.e., \algname-NR, which solves the optimization problem \eqref{eq:coarse_registration} in closed-form without RANSAC, and \algname-R, which utilizes RANSAC for coarse registration---to the GSplat registration methods GaussReg \cite{chang2025gaussreg} and PhotoReg \cite{yuan2024photoreg}, in addition to RANSAC-based global registration (RANSAC-GR) \cite{fischler1981random, holz2015registration}, Fast Global Registration (FGR) \cite{zhou2016fast}, and variants of the Iterative Closest Point (ICP) \cite{rusinkiewicz2001efficient, park2017colored}. We evaluate each method not only on standard benchmark datasets for radiance fields, but also on real-world data collected by heterogeneous robot platforms, including a quadruped, drone, and manipulator (in the case of SIREN). In all our experiments, we only require the trained GSplat models as input; however, some of the baselines require access to the set of camera poses, which we provide when evaluating these methods.
Further, we ablate the different components of \algname, to quantify the relative improvements in performance provided by each component, and examine the gains in visual fidelity afforded by finetuning the fused model. We provide these results in Appendix~\ref{sec:appendix_experiments}, as well as additional discussion of the results presented in this section. 
Lastly, we demonstrate \algname in collaborative multi-robot mapping, where the mapping task cannot be accomplished by a single robot, necessitating mapping with multiple robots for task success. 

\phantomsection
\label{ssec:experiment_metrics}
\smallskip
\noindent\textbf{Experimental Setup and Metrics.}
For the real-world robot data, we utilize the Unitree Go1 Quadruped and a Modal AI drone with an onboard camera and the Franka Panda manipulator with a wrist camera to collect RGB images. In addition, we evaluate all methods on the real-world scenes in the Mip-NeRF360 dataset \cite{barron2022mip}, a state-of-the-art benchmark dataset for neural rendering. We train the GSplat models using the original implementation provided by the authors of \cite{kerbl20233d} for baselines which require this pipeline and utilize Nerfstudio \cite{tancik2023nerfstudio} for \algname. We execute SIREN on a desktop computer with a 24GB NVIDIA GeForce RTX 3090 GPU and the baselines on an H20 GPU after training the GSplat maps for $30000$ iterations.
We note that in robotics, the geometric fidelity of robot's map is of significant importance for effective localization and collision avoidance. Hence, we compare all methods in terms of the rotation error (RE) [deg.], translation error (TE), and scale error (SE) [in non-metric units] attained by each method, in addition to the computation time (CT) [sec.]. Moreover, we examine the photometric quality of the fused maps generated by each method, computing the peak signal-to-noise ratio (PSNR), the structural similarity index measure (SSIM), and the learned perceptual image patch similarity (LPIPS), standard metrics in the computer vision community for assessing visual fidelity. 
We provide color-coded results for each metric with the red shade denoting the top-performing statistic, the yellow shade denoting the second-best, and the green shade denoting the third-best. In all the registration methods, we do not pre-process the individual submaps to remove floaters (i.e., non-existing geometry). Consequently, floaters present in these submaps are retained in the fused map.

\subsection{Mip-NeRF360 Dataset}
We utilize the \emph{Playroom}, \emph{Truck}, and \emph{Room} scenes in the Mip-NeRF360 Dataset. These real-world scenes were all collected in realistic settings with natural lighting effects, both indoors and outdoors. While the \emph{Playroom} and \emph{Room} scenes were captured indoors, the \emph{Truck} scene was captured outdoors. We split the datasets into two subsets with varying overlap. Specifically, the first subset of the \emph{Truck} scene captures the left side of the truck, while the second subset captures the right side of the truck. The only overlap between both subsets occurs at the front and rear of the truck. We split the \emph{Room} scene into two subsets following the same procedure. In the \emph{Playroom} scene, we allow for greater overlap, with the density of images per subregion of the scene varying between both subsets. We train independent GSplat maps for each scene-subset pair.  

\smallskip
\noindent\textbf{Geometric Evaluation.}
In \Cref{tab:baseline_geometric_performance_metrics}, we report the geometric errors of each registration method across the three scenes. \algname-R, our method, achieves the lowest rotation and translation errors in two of the three scenes (\emph{Playroom} and \emph{Truck}): with about $1.14$x to $8.89$x lower rotation errors and about $6$x to $46$x lower translation errors compared to the baseline methods. Meanwhile, in the \emph{Room} scene, \algname-NR achieves the lowest translation and scale error, with \algname-R achieving the second-best performance on these metrics. In summary, \algname achieves the lowest geometric errors (i.e., rotation, translation, and scale errors) across all scenes, except the rotation error in the \emph{Room} scene.

\begin{table*}[th]
	\centering
	\caption{Geometric performance of the registration algorithms on the Mip-NeRF360 dataset (see Section~\ref{ssec:experiment_metrics} for a description of the metrics).}
	\label{tab:baseline_geometric_performance_metrics}
	\begin{adjustbox}{width=\linewidth}
		{\begin{tabular}{l | c c c c | c c c c | c c c c}
				\toprule
                    & \multicolumn{4}{c |}{\emph{Playroom}} & \multicolumn{4}{c |}{\emph{Truck}} & \multicolumn{4}{c}{\emph{Room}} \\
				Methods & RE $\downarrow$ & TE  $\downarrow$ & SE $\downarrow$ & CT $\downarrow$ & RE $\downarrow$ & TE $\downarrow$ & SE $\downarrow$ & CT $\downarrow$ & RE $\downarrow$ & TE $\downarrow$ & SE $\downarrow$ & CT $\downarrow$ \\
				\midrule
                    PhotoReg \cite{yuan2024photoreg} & 6.036 & 18806 & 841.3 & 2177 & 177.3 & 2856 & 444.0 & 1814 & \cellcolor{WildStrawberry!40}0.161 & 4983 & 452.7 & 1409 \ \\
                    GaussReg \cite{chang2025gaussreg} & 0.766  & 55.50 & \cellcolor{GreenYellow!40}0.364 & 15.06 & 21.10 & \cellcolor{GreenYellow!40}316.3 & 16.76 & 5.174 & 7.464  & 628.3 & \cellcolor{GreenYellow!40}91.97 & 6.932 \\
                    RANSAC-GR \cite{fischler1981random, holz2015registration} & 4.835 & 56.22 & 17.85 & 0.996 & 46.72 & 2642 & \cellcolor{GreenYellow!40}13.64 & \cellcolor{WildStrawberry!40}{2.569} & 8.139 & \cellcolor{GreenYellow!40}194.7 & 152.5 & 0.517 \\
                    FGR \cite{zhou2016fast} & 2.988 & 18.83 & 14.37 & \cellcolor{WildStrawberry!40}{0.887} & 3.778 & 2231 & 79.45 & 3.480 & 4.869 & 265.6 & 219.6 & \cellcolor{WildStrawberry!40}{0.511} \\
                    ICP \cite{rusinkiewicz2001efficient} & 2.362 & 19.11 & 14.37 & 2.127 & \cellcolor{GreenYellow!40}3.672 & 2232 & 79.45 & 3.805 & 5.154 & 266.1 & 219.6 & 1.579 \\
                    Colored-ICP \cite{park2017colored} & \cellcolor{Goldenrod!40}{0.194} & \cellcolor{GreenYellow!40}{12.28} & \textcolor{black}{14.37} & 3.951 & 4.043 & 2250 & 79.45 & 6.392 & 2.256 & 232.7 & 219.6 & 3.815 \\
                    \algname-NR [{Ours}] & \cellcolor{GreenYellow!40}0.348 & \cellcolor{Goldenrod!40}{4.860} & \cellcolor{Goldenrod!40}{0.282} & 41.16 & \cellcolor{Goldenrod!40}{0.511} & \cellcolor{Goldenrod!40}{8.07} & \cellcolor{Goldenrod!40}9.581 & 53.42 & \cellcolor{GreenYellow!40}{0.381} & \cellcolor{WildStrawberry!40}{2.648} & \cellcolor{WildStrawberry!40}{1.016} & 40.24 \\
                    \algname-R [{Ours}] & \cellcolor{WildStrawberry!40}{0.170} & \cellcolor{WildStrawberry!40}{1.933} & \cellcolor{WildStrawberry!40}{0.170} & 39.73 & \cellcolor{WildStrawberry!40}{0.413} & \cellcolor{WildStrawberry!40}{6.845} & \cellcolor{WildStrawberry!40}{2.548} & 52.47 & \cellcolor{Goldenrod!40}{0.237} & \cellcolor{Goldenrod!40}{3.289} & \cellcolor{Goldenrod!40}{2.673} & 39.71 \\
				\bottomrule
		\end{tabular}}
	\end{adjustbox}
\end{table*}

\begin{figure}[th]
    \centering
    \includegraphics[width=\linewidth]{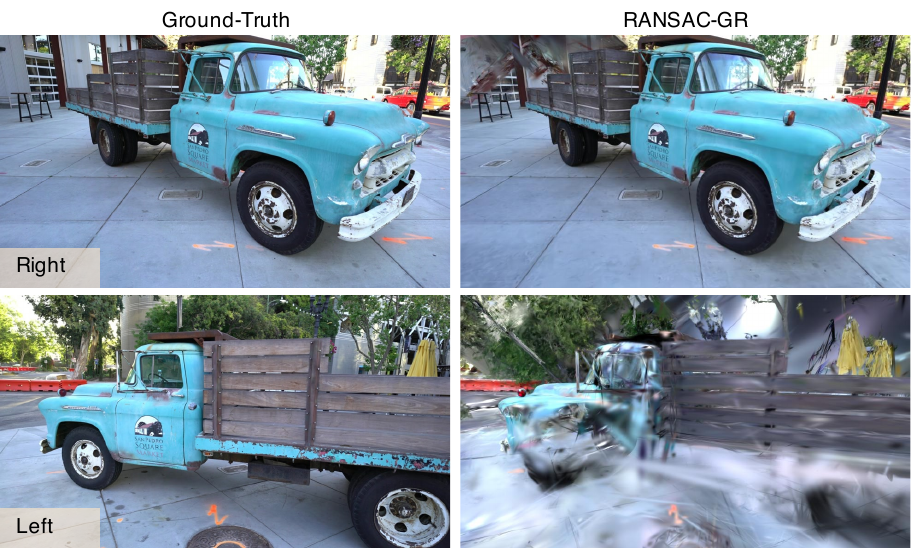}
    \caption{Although RANSAC-GR achieves the highest mean PSNR and SSIM scores and the lowest LPIPS score in the \emph{Truck} scene, RANSAC-GR does not accurately register the individual GSplat maps. While the right side of the truck in the RANSAC-GR fused map looks similar to the ground-truth image (shown in the top panel), the left side of the truck is missing (shown in the bottom panel). The standard deviation of the PSNR, SSIM, and LPIPS scores achieved by RANSAC-GR reflects the actual registration performance of the method.}
    \label{fig:ransac_gr_vs_ground_truth}
\end{figure}

\begin{figure*}[th]
    \centering
    \includegraphics[width=\linewidth]{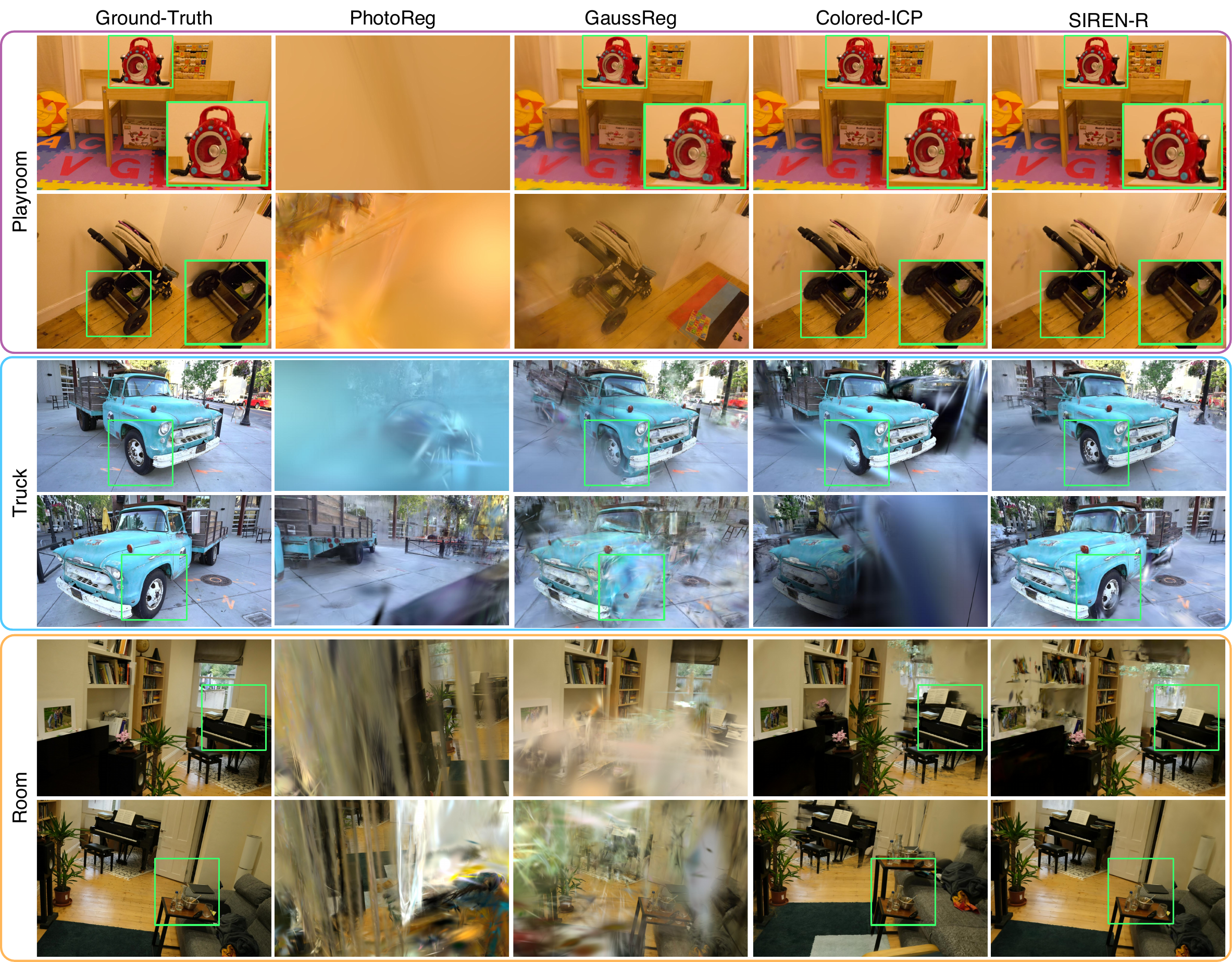}
    \caption{Rendered images from the fused GSplat maps of the \emph{Playroom}, \emph{Truck}, and \emph{Room} scenes. \algname generates high-fidelity fused GSplat maps, evidenced by the precise geometric detail in the images, visible in the regions indicated by the green squares. Inaccurate registration of GSplat maps generally result in artifacts in the rendered images.}
    \label{fig:photometric_performance_rendered_images}
\end{figure*}

\smallskip
\noindent\textbf{Photometric Evaluation.}
Now, we examine the photometric performance of the GSplat registration methods reported in \Cref{tab:baseline_photometric_performance_metrics} in Appendix~\ref{sec:appendix_experiments}. \algname-R achieves the best photometric performance in the \emph{Playroom} scene, with the highest mean PSNR and SSIM and lowest mean LPIPS scores. Similarly, in the \emph{Room} scene, \algname-NR achieves the best photometric performance across all metrics, followed by \algname-R. In the \emph{Truck} scene, RANSAC-GR achieves the best mean PSNR and SSIM scores. Although this finding may appear inconsistent with the geometric results presented in \Cref{tab:baseline_geometric_performance_metrics}, the high standard deviation of each of the scores achieved by RANSAC-GR (about $2$x to $3$x larger than that of \algname) suggests that the geometric and photometric performance metrics for this scene might be consistent, indicating that the fused map generated by RANSAC-GR warrants further examination.
We provide rendered images from the fused map generated by RANSAC-GR compared to the ground-truth images in \Cref{fig:ransac_gr_vs_ground_truth} to examine the registration results of RANSAC-GR. From \Cref{fig:ransac_gr_vs_ground_truth}, we note that RANSAC-GR fails to accurately register the left and right sides of the truck. In fact, the left side of the truck is missing in the bottom panel associated with RANSAC-GR in \Cref{fig:ransac_gr_vs_ground_truth}. However, this failure mode is not fully captured by the mean score of the photometric performance metrics, since the rendered images of the right side of the truck (shown in the top panel in \Cref{fig:ransac_gr_vs_ground_truth}) look quite similar to the corresponding ground-truth images. In conclusion, RANSAC-GR does not accurately register the individual GSplat maps, despite achieving the highest mean PSNR and SSIM scores in the \emph{Truck} scene.
In \Cref{fig:photometric_performance_rendered_images}, we show the rendered images from the fused GSplat maps generated by the registration methods from different viewpoints compared to the ground-truth images. We visualize a pair of images from the \emph{Playroom}, \emph{Truck}, and \emph{Room} scenes, restricting our visualizations to PhotoReg, GaussReg, Colored-ICP, and \algname-R due to space considerations.

\subsection{Mobile-Robot Mapping}
\begin{figure*}[th]
    \centering
    \includegraphics[width=\linewidth]{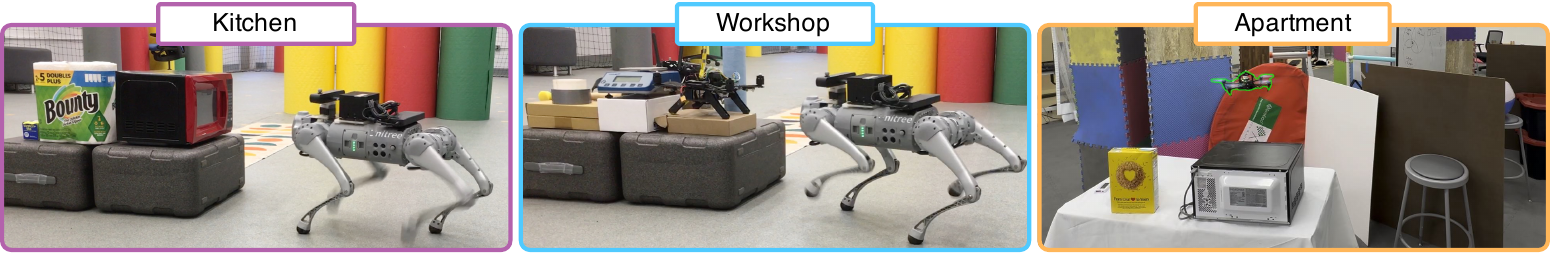}
    \caption{Stillshots of a quadruped mapping different areas of a kitchen and workshop and a drone mapping an apartment-like scene. Each robot trains independent GSplat submaps of the areas it mapped. The submaps of each scene are registered to obtain a composite map covering the entirety of the scene.}
    \label{fig:photometric_performance_mobile_robot_mapping}
\end{figure*}

We utilize a quadruped and a drone to map three environments, depicted in \Cref{fig:photometric_performance_mobile_robot_mapping}. The quadruped maps the \emph{Kitchen} and \emph{Workshop} environments, while the drone maps an \emph{Apartment} scene, with multiple partitioned room-like areas. The robots create submaps in each environment individually, containing different regions of the scene. The submaps in the \emph{Kitchen} and \emph{Workshop} scenes have minimal overlap, while the submaps in the \emph{Apartment} scene have greater overlap. Since each submap is trained independently in different reference frames, fusing the submaps requires registration of the maps. Here, we examine the performance of GaussReg, PhotoReg, and two variants of \algname: \algname-NR and \algname-R, in registering the submaps in each scene to obtain a composite map of the entire scene.

\smallskip
\noindent\textbf{Geometric Performance.}
\Cref{tab:baseline_geometric_performance_metrics_mobile_robot_mapping} summarizes the geometric errors of each algorithm, showing that \algname achieves the best geometric performance across all scenes, with the top-two-performing methods being the variants of \algname. Specifically, in the \emph{Kitchen} scene, \algname-NR achieves the lowest rotation, translation, and scale errors by a factor of about $160$x, $465$x, and $488$x, respectively, compared to the best-performing baseline. The performance of \algname-R closely follows that of \algname-NR. Similarly, in the \emph{Workshop} scene, \algname-R achieves the lowest rotation and translation errors by a factor of $415$x and $1287$x, respectively, compared to the best-performing baseline, followed by \algname-NR, while \algname-NR achieves the lowest scale error by a factor of $2962$x, followed by \algname-R. Lastly, \algname-R achieves the lowest rotation, translation, and scale errors in the \emph{Apartment} scene, followed by \algname-NR. GaussReg requires the least computation time across all scenes, while PhotoReg requires the greatest computation time. Although compared to GaussReg \algname requires a notably greater computation time, \algname requires much lower computation times compared to PhotoReg.

\begin{figure*}[th]
    \centering
    \includegraphics[width=0.85\linewidth]{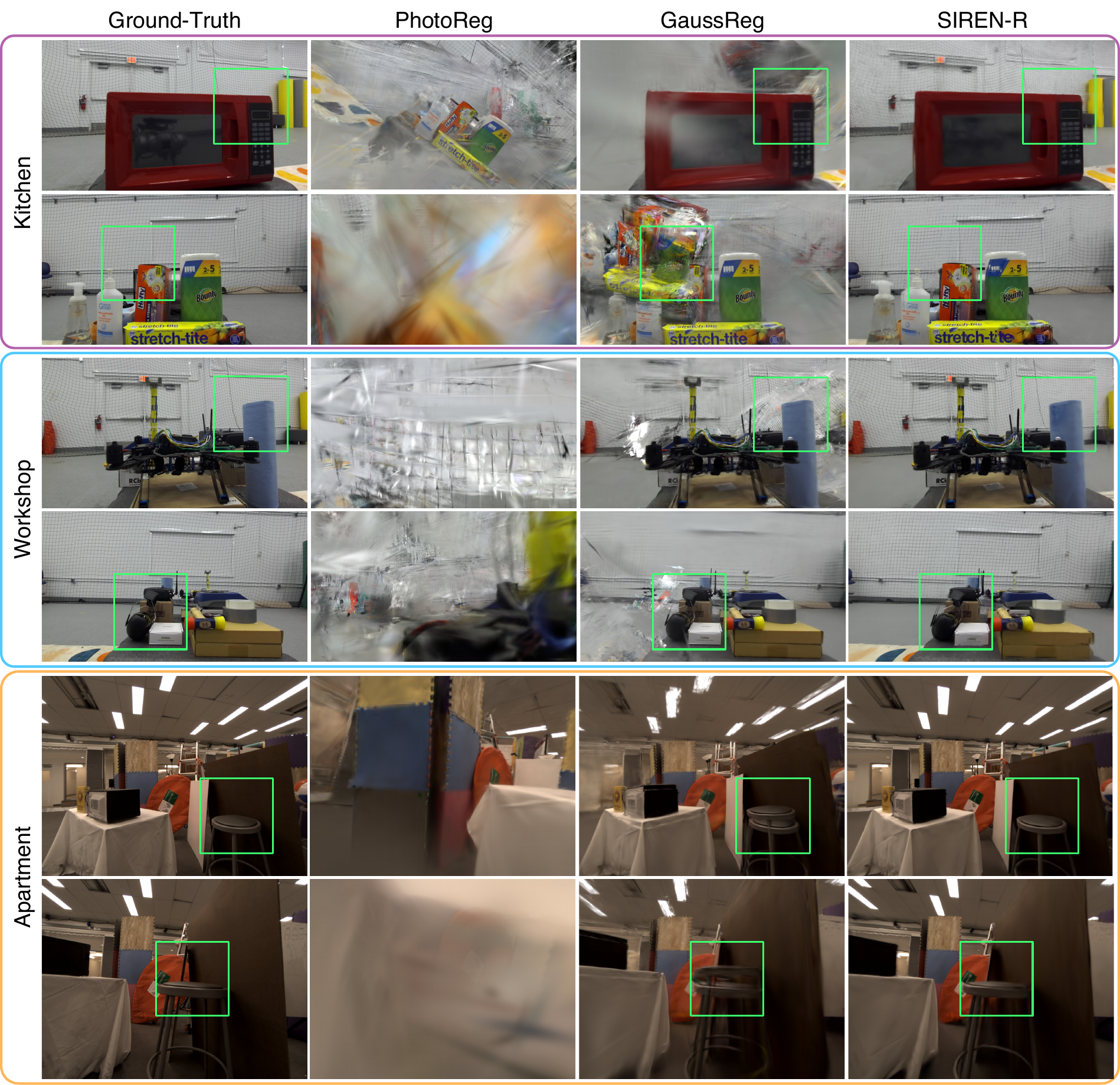}
    \caption{Rendered images from the fused GSplat maps of the \emph{Kitchen}, \emph{Workshop}, and \emph{Apartment} scenes mapped by a quadruped and drone. Unlike other competing methods, \algname generates fused GSplat maps of high visual fidelity, e.g., in the regions indicated by the green squares.}
    \label{fig:photometric_performance_mobile_robot_rendered_images}
\end{figure*}

\smallskip
\noindent\textbf{Photometric Performance.}
Further, we examine the photometric quality of the fused map generated by the GSplat registration methods across the three scenes. In line with the geometric results, \algname outperforms all the baseline methods, as reported in \Cref{tab:baseline_photometric_performance_metrics_mobile_robot_mapping}. While \algname-R achieves the best photometric scores (i.e., the highest PSNR and SSIM scores and lowest LPIPS scores) in the \emph{Workshop} scene, \algname-NR attains the best-performing PSNR, SSIM, and LPIPS scores in the \emph{Kitchen} scene, followed by \algname-R. In the \emph{Apartment} scene, \algname-R achieves the best PSNR score and LPIPS (tied with \algname-NR), while \algname-NR also achieves the best SSIM score. GaussReg outperforms PhotoReg in all scenes.
In addition to the results in \Cref{tab:baseline_photometric_performance_metrics_mobile_robot_mapping}, we provide rendered images from each of the fused map in \Cref{fig:photometric_performance_mobile_robot_rendered_images} for qualitative evaluation of the performance of each method. 

\begin{table*}[th]
	\centering
	\caption{Geometric performance of GSplat registration algorithms in mobile-robot mapping.}
	\label{tab:baseline_geometric_performance_metrics_mobile_robot_mapping}
	\begin{adjustbox}{width=\linewidth}
		{\begin{tabular}{l | c c c c | c c c c | c c c c}
				\toprule
                    & \multicolumn{4}{c |}{\emph{Kitchen}} & \multicolumn{4}{c |}{\emph{Workshop}} & \multicolumn{4}{c}{\emph{Apartment}} \\
				Methods & RE $\downarrow$ & TE  $\downarrow$ & SE $\downarrow$ & CT $\downarrow$ & RE $\downarrow$ & TE $\downarrow$ & SE $\downarrow$ & CT $\downarrow$ & RE $\downarrow$ & TE $\downarrow$ & SE $\downarrow$ & CT $\downarrow$ \\
				\midrule
                    PhotoReg \cite{yuan2024photoreg} & \cellcolor{GreenYellow!40}40.49 & 2350 & 413.37 & 1042 &  140.5 & 10052 & 4310 & 934.2 & 24.09 & 4433 & 260.2 & 801.0 \\
                    GaussReg \cite{chang2025gaussreg} & 40.89 & \cellcolor{GreenYellow!40}1477 & \cellcolor{GreenYellow!40}171.8 & \cellcolor{WildStrawberry!40}11.33 & \cellcolor{GreenYellow!40}55.66 & \cellcolor{GreenYellow!40}9531 & \cellcolor{GreenYellow!40}4305 &  \cellcolor{WildStrawberry!40}5.491 & \cellcolor{GreenYellow!40}3.114 & \cellcolor{GreenYellow!40}102.6 & \cellcolor{GreenYellow!40}13.59 & \cellcolor{WildStrawberry!40}5.4983 \\
                    \algname-NR [\textbf{Ours}] & \cellcolor{WildStrawberry!40}0.253 & \cellcolor{WildStrawberry!40}3.173 & \cellcolor{WildStrawberry!40}0.352 & 59.22 & \cellcolor{Goldenrod!40}0.518 & \cellcolor{Goldenrod!40}11.77 & \cellcolor{WildStrawberry!40}1.453 & 67.98 & \cellcolor{Goldenrod!40}0.148 & \cellcolor{Goldenrod!40}1.758 & \cellcolor{Goldenrod!40}0.605 & 35.91 \\
                    \algname-R [\textbf{Ours}] & \cellcolor{Goldenrod!40}0.430 & \cellcolor{Goldenrod!40}4.795 & \cellcolor{Goldenrod!40}3.849 & 56.14 & \cellcolor{WildStrawberry!40}0.134  & \cellcolor{WildStrawberry!40}7.400 & \cellcolor{Goldenrod!40}10.88 & 55.16 & \cellcolor{WildStrawberry!40}0.119 & \cellcolor{WildStrawberry!40}1.495 & \cellcolor{WildStrawberry!40}0.102 & 34.22 \\
				\bottomrule
		\end{tabular}}
	\end{adjustbox}
\end{table*}

\subsection{Tabletop Mapping with Multiple Manipulators}
\begin{figure*}[th]
    \centering
    \includegraphics[width=\linewidth]{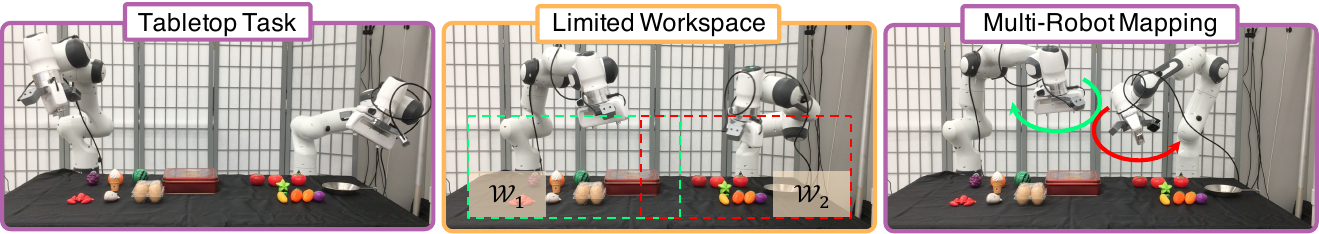}
    \caption{Tabletop robotics tasks, e.g., manipulation, generally require robots to map the scene prior to completing the task. However, the limited workspace of each robot often demands assistance from a human-operator or improvised hardware, e.g., selfie sticks. \algname eliminates these challenges, via registration of the local maps trained by each robot to construct a global map consistent with the real-world.}
    \label{fig:photometric_performance_manipulation_scene}
\end{figure*}

\begin{figure*}[th]
    \centering
    \includegraphics[width=0.85\linewidth]{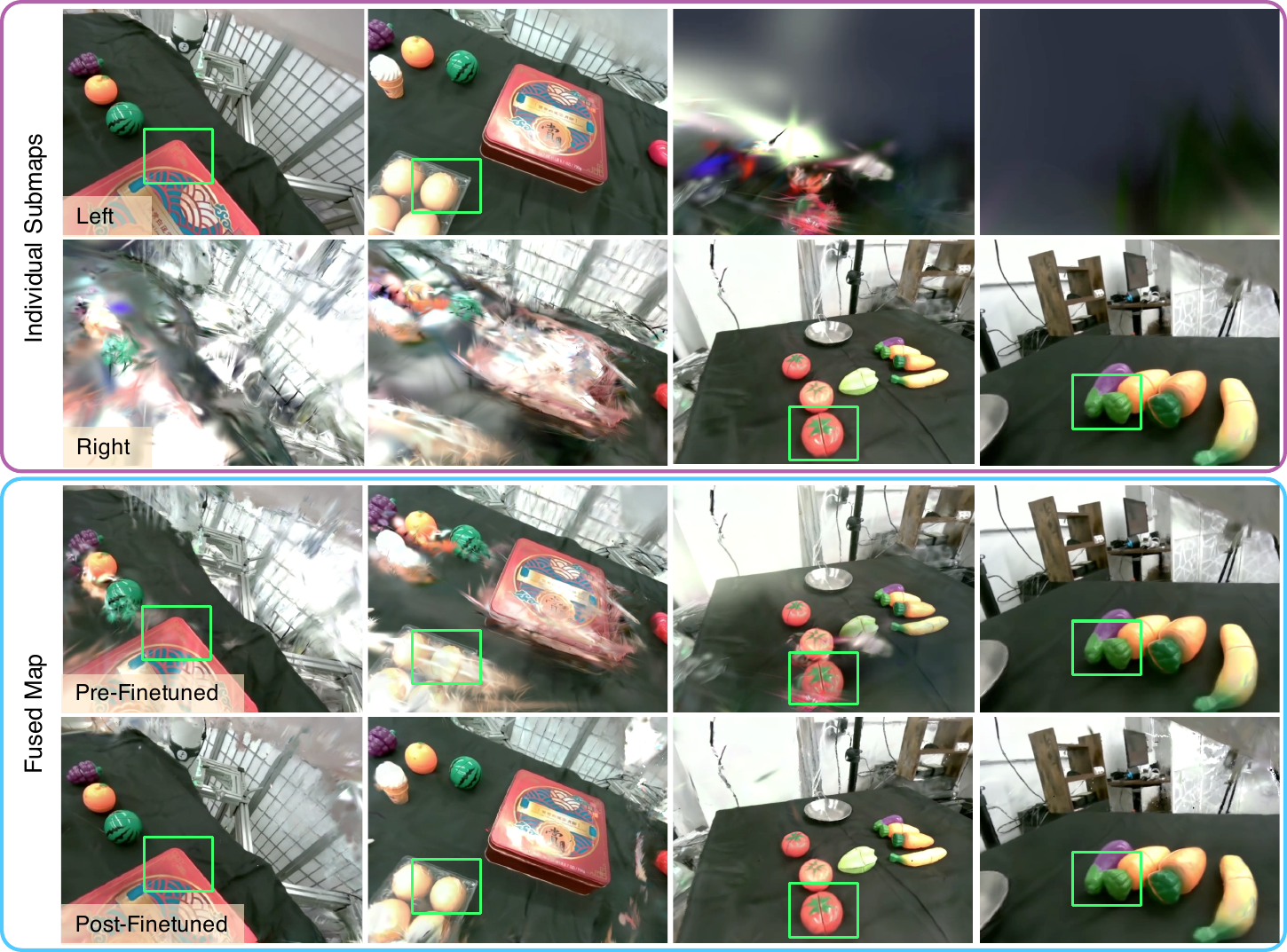}
    \caption{Rendered images of the local maps of a tabletop scene trained by two manipulators. The maps provide high-fidelity reconstructions within the workspace of each robot, but fail to represent the real-world in regions outside the workspace. \algname fuses the local maps to generate a high-fidelity global map consistent with the entirety of the scene, especially after finetuning on data rendered directly from the GSplat to remove floaters, without any interaction with the real-world, as indicated by the green squares.}
    \label{fig:photometric_performance_manipulation_maps}
\end{figure*}

We demonstrate the effectiveness of \algname in tabletop robotics tasks with fixed-base manipulators, which often require the robots to map the scene prior to the task, e.g., in manipulation \cite{shen2023distilled, shorinwa2024splat}. In \Cref{fig:photometric_performance_manipulation_scene}, we provide an example with two Franka robots, each with a wrist camera. Due to the limited workspace of each robot, visualized in \Cref{fig:photometric_performance_manipulation_scene}, mapping often requires the assistance of a human-operator \cite{shorinwa2024splat} or ad-hoc solutions such as hardware improvisation, e.g., using selfie sticks \cite{shen2023distilled}. By enabling the fusion of GSplat maps trained individually by each robot, \algname effectively eliminates these limitations. In other words, with \algname, each robot can train a submap within its reachable workspace and still recover the global map via registration with \algname. In \Cref{fig:photometric_performance_manipulation_maps}, we show the submaps trained by each robot. As expected, each robot has a high-fidelity submap within the confines of its reachable workspace, evident in the first-two images in the left robot's map and the last-two images in the right's robot map in \Cref{fig:photometric_performance_manipulation_maps}. In areas outside of its reachable workspace, the robot's map fails to represent the real world accurately, visible in the last-two images in the left robot's map and the first-two images in the right's robot map. With \algname, each robot obtains a higher-fidelity map over a much broader region of the environment. However, floaters present in the submaps can degrade the quality of the fused map in certain regions. To address this challenge, we finetune the fused map for about $70.98$ secs using images generated entirely from the GSplat maps, i.e., we do not require any real-world data. We provide rendered images from the finetuned fused map in \Cref{fig:photometric_performance_manipulation_maps}, showing near-perfect reconstruction of the global scene. We explore the finetuning procedure in Appendix~\ref{ssec:finetuning}.

\section{Conclusion}
\label{sec:conclusion}
We present \algname, a semantics-grounded registration algorithm for multi-robot GSplat maps that neither requires access to camera poses or images nor initialization of inter-map relative transforms. \algname harnesses the robustness of semantics to: (a) identify candidate matches between Gaussians across the input maps, (b) compute a coarse transformation aligning both maps from a Gaussian-to-Gaussian registration problem posed as an optimization program, and (c) refine the coarse registration result for high-accuracy fusion of local submaps into a high-fidelity global map. We demonstrate the versatility of \algname across maps constructed by robots of different embodiments, including a quadruped, drone, and manipulator, highlighting the superior performance of \algname compared to GSplat registration algorithms and classical point-cloud registration methods.

\section{Limitations and Future Work}
\label{sec:limitations_future_work}
Our registration algorithm relies on semantics for robust, initialization-free registration, and thus requires that the input maps have embedded semantic codes. A GSplat map may lack semantic information if the map was not trained with semantics or if the scene lacks any semantically-relevant features, which would be a tail event in practical situations. We can post-train GSplats to embed semantics in $3$D into the map or leverage $2$D vision foundation models to directly extract semantic information from RGB images rendered from the GSplat by back-projecting $2$D pixels into the $3$D world.
Radiance fields are prone to generate floaters in areas of the scene with little to no supervision, which can degrade the fidelity of the map. The resulting floaters are retained in the fused map, which could ultimately reduce the accuracy of the map. However, by finetuning the fused map with synthetic data, i.e., images rendered from the map as opposed to real-world images, floaters in the map can be removed for high-fidelity mapping.

\section*{Acknowledgments}
This work was supported in part by NSF grant 2342246, NSF CAREER Award 2044149, Office of Naval Research N00014-23-1-2148, and Princeton SEAS Innovation Award from The Addy Fund for Excellence in Engineering. Toyota Research Institute provided funds to support this work. Jiankai Sun is partially supported by Stanford Interdisciplinary Graduate Fellowship. 

\clearpage

\bibliographystyle{plainnat}
\bibliography{references}

\clearpage

\begin{appendices}
    \section{Coarse Gaussian-to-Gaussian Registration}
\label{app:method}
We discuss the derivation of the closed-form solution to \eqref{eq:coarse_registration}.
Let the objective function of \eqref{eq:coarse_registration} be denoted by $J$.
From the first-order optimality conditions: 
\begin{equation}
    \nabla_{t} J = \sum_{(i, j) \in \mcal{E}} \left( w_{ij} s_{c}Rp_{i} + t - q_{j} \right) = 0,
\end{equation}
yielding the optimal translation:
\begin{equation}
    \label{eq:coarse_registration_opt_translation_app}
    t_{c}^{\star} = \tilde{\mu}_{\mcal{Q}} - s_{c}^{\star} R^{\star} \tilde{\mu}_{\mcal{P}},
\end{equation}
where ${\tilde{\mu}_{\mcal{P}}}$ and ${\tilde{\mu}_{\mcal{Q}}}$ denote the weighted average of the means of the Gaussians in $\mcal{P}$ and $\mcal{Q}$, with weights $w_{ij}$ for Gaussian $i$ in $\mcal{P}$ and Gaussian $j$ in $\mcal{Q}$.
By substituting the optimal value of $t$ \eqref{eq:coarse_registration_opt_translation_app} in \eqref{eq:coarse_registration}, we obtain the following optimization problem over $s_{c}$ and $R$:
\begin{equation}
    \label{eq:coarse_registration_frob}
    \begin{aligned}
        \minimize{s_{c} \in \mbb{R}_{++}, R \in \SO(3)} &\frac{1}{2} \norm{(s_{c}R\check{P} - \check{Q})W^{\frac{1}{2}}}_{F}^{2} \\
        & + \frac{1}{2} \sum_{(i, j) \in \mcal{E}}  w_{ij} \norm{s_{c}R \check{H}_{p_{i}} - \check{H}_{q_{j}}}_{F}^{2},
    \end{aligned}
\end{equation}
where ${\check{P} \in \mbb{R}^{3 \times N}}$ and ${\check{Q} \in \mbb{R}^{3 \times N}}$ represent the \emph{zero-centered} Gaussians in $\mcal{P}$ and ${\mcal{Q}}$, respectively, with the $i$th column of $\check{P}$ given by ${\check{P}_{i} = p_{i} - \tilde{\mu}_{\mcal{P}}}$ and similarly for the $j$th column of $\check{Q}$, and 
${\check{H}_{p_{i}} = H_{p_{i}}  \Lambda_{p_{i}}}$ and ${\check{H}_{q_{j}} = H_{q_{j}}  \Lambda_{q_{j}}}$. Lastly, ${W \in \mbb{R}^{N \times N}}$ denotes the diagonal weight matrix, ${W_{kk} = w_{k}}$, with ${w_{k} = w_{ij}}$,~${\forall k = (i, j) \in \mcal{E}}$.
Now, we can reformulate \eqref{eq:coarse_registration_frob} as a trace-minimization problem, by leveraging the relation: ${\norm{A}_{F}^{2} = \trace(A^{\top}A)}$ for any real-valued matrix ${A \in \mbb{R}^{m \times n}}$.
Reformulating the problem as a trace-minimization problem enables us to decompose the norm-minimization problem \eqref{eq:coarse_registration_frob} into a nested pair of subproblems: an outer subproblem over $s_{c}$ and an inner subproblem over $R$. We can simplify the inner subproblem into the form:
\begin{equation}
    \label{eq:coarse_registration_trace_R}
    \begin{aligned}
        \minimize{R \in \SO(3)} &-\trace\left(R^{\top} \left(\check{Q}W\check{P}^{\top}
        + \sum_{(i, j) \in \mcal{E}} w_{ij} \check{H}_{q_{j}}\check{H}_{p_{i}}^{\top}\right)\right),
    \end{aligned}
\end{equation}
which affords a closed-form optimal solution, with 
\begin{equation}
    \label{eq:coarse_registration_opt_R_app}
    R_{c}^{\star} = U_{c} \Theta_{c} V_{c}^{\top},
\end{equation}
where ${U_{c} \Sigma_{c} V_{c}^{\top} = \check{Q}W\check{P}^{\top}
+ \sum_{(i, j) \in \mcal{E}} w_{ij} \check{H}_{q_{j}}\check{H}_{p_{i}}^{\top}}$, computed via the singular value decomposition (SVD),
and ${\Theta_{c} = \diag(1, 1, \det(U_{c}V_{c}^{\top}))}$. Using the first-order optimality condition, we can compute the optimal scale after computing the optimal rotation from the outer subproblem given by:
\begin{equation}
    \label{eq:coarse_registration_trace_scale}
    \begin{aligned}
        \minimize{s_{c} \in \mbb{R}_{++}} & \frac{1}{2} s_{c}^{2} \trace\left(W\check{P}^{\top}\check{P} + \sum_{(i, j) \in \mcal{E}} w_{ij} \check{H}_{p_{i}}^{\top}\check{H}_{p_{i}}\right) \\
        & \hspace{-1em} - s_{c}\trace\left(R^{\top} \left(\check{Q} W \check{P}^{\top} + \sum_{(i, j) \in \mcal{E}} w_{ij} \check{H}_{q_{j}}\check{H}_{p_{i}}^{\top} \right) \right),
    \end{aligned}
\end{equation}
yielding the optimal solution:
\begin{equation}
    \label{eq:coarse_registration_opt_scale_app}
    s_{c}^{\star} = \frac{\trace(\Theta_{c} \Sigma)}{\trace\left(W\check{P}^{\top}\check{P} + \sum_{(i, j) \in \mcal{E}} w_{ij} \check{H}_{p_{i}}^{\top}\check{H}_{p_{i}}\right)}.
\end{equation}
For brevity, we omit further analysis of the optimality of the solution and refer interested readers to \cite{umeyama1991least} for the proof of a related problem, which applies to the problem considered in this work.

\section{Experiments}
\label{sec:appendix_experiments}
We report the photometric performance of each registration method with the Mip-NeRF360 dataset and the data collected by the robots in our experiments and provide further discussion of the experimental results. In addition, we present ablations, examining the different components of \algname. Lastly, we explore finetuning the resulting composite maps for higher visual fidelity.

\subsection{Mip-NeRF360 Dataset}
\label{ssec:appendix_experiments_mip_nerf_360}
\smallskip
\noindent\textbf{Geometric Evaluation.}
In the \emph{Room} scene, PhotoReg achieves the lowest rotation error by a factor of about $1.47$x but also achieves the largest translation and scale errors. 
Based on the results across all scenes, \algname almost always consistently outperforms competing methods. From \Cref{tab:baseline_geometric_performance_metrics}, RANSAC-GR and FGR achieve the fastest computation times; however, RANSAC-GR and FGR do not generally achieve consistently low geometric errors. Although \algname is slower than the classical point-cloud registration algorithms and GaussReg, \algname generally outperforms these methods in accuracy by significant margins. Moreover, about $40\%$ to $50\%$ of the total computation time of \algname is spent on the semantics extraction procedure. Hence, the total computation time can be significantly improved by utilizing faster semantics distillation methods, e.g., \cite{shorinwa2024fast}.

\smallskip
\noindent\textbf{Photometric Evaluation.}
From \Cref{fig:photometric_performance_rendered_images}, in the \emph{Playroom} scene, PhotoReg fails to sufficiently register the individual maps to obtain photorealistic renderings. In contrast, GaussReg, Colored-ICP, and \algname-R generate high-fidelity renderings. However, the fused maps generated by GaussReg and Colored-ICP are not accurately aligned, compared to that of \algname-R, as evidenced in insets in the images. The fused map in GaussReg and Colored-ICP contain duplicate objects due to inaccurate registration of the individual maps. In contrast, \algname-R provides greater accuracy. Likewise, \algname-R achieves the highest-fidelity rendering in the \emph{Truck} scene with consistent geometry, whereas Colored-ICP fails to register the left and right sides of the truck. Although GaussReg fuses both sides of the truck, GaussReg fails to compute a high-accuracy transform, resulting in the artifacts visible in \Cref{fig:photometric_performance_rendered_images}. Although PhotoReg registers the cargo bed of the truck in both maps, PhotoReg fails to align the truck accurately in terms of the rotation transform, with the front end of the truck in one map registered to the rear end of the truck in the other map.
Finally, in the \emph{Room} scene, whereas \algname-R generates high-fidelity rendered images, other methods fail to accurately register the individual maps. In particular, Colored-ICP generates a fused map with duplicate objects, e.g., the piano and the table, indicated by the green squares, while PhotoReg and GaussReg generate fused maps with notable artifacts.

\begin{table*}[th]
	\centering
	\caption{Photometric performance of registration algorithms for GSplat maps from the Mip-NeRF360 dataset.}
	\label{tab:baseline_photometric_performance_metrics}
	\begin{adjustbox}{width=\linewidth}
		{\begin{tabular}{l | c c c | c c c | c c c }
				\toprule
                    & \multicolumn{3}{c |}{\emph{Playroom}} & \multicolumn{3}{c |}{\emph{Truck}} & \multicolumn{3}{c}{\emph{Room}} \\
				Methods & PSNR $\uparrow$ & SSIM  $\uparrow$ & LPIPS  $\downarrow$ & PSNR $\uparrow$ & SSIM  $\uparrow$ & LPIPS  $\downarrow$ & PSNR $\uparrow$ & SSIM  $\uparrow$ & LPIPS  $\downarrow$ \\
				\midrule
                    PhotoReg \cite{yuan2024photoreg} & 11.5 $\pm$ 2.3 &   0.68 $\pm$ 0.11 & 0.67 $\pm$ 0.12 & 10.6 $\pm$ 0.9 & 0.39 $\pm$ 0.07 & 0.72 $\pm$ 0.08 & 10.2 $\pm$ 1.2 & 0.46 $\pm$ 0.07 &  0.78 $\pm$  0.04 \\
                    GaussReg \cite{chang2025gaussreg} & 23.7 $\pm$ 3.3 &  0.86 $\pm$ 0.06 & 0.22 $\pm$ 0.08 & 13.4 $\pm$ 1.9 & 0.54 $\pm$ 0.12 & 0.53  $\pm$ 0.13  &  13.4 $\pm$ 3.6  &  0.61 $\pm$ 0.11  &  0.55  $\pm$  0.15 \\
                    RANSAC-GR \cite{fischler1981random, holz2015registration} & 17.9 $\pm$ 3.2 & 0.77 $\pm$ 0.09 & 0.37 $\pm$ 0.09 & \cellcolor{WildStrawberry!40}{18.9 $\pm$ 7.0} & \cellcolor{WildStrawberry!40}{0.66 $\pm$ 0.24} & \cellcolor{Goldenrod!40}{0.32 $\pm$ 0.22} & 14.2 $\pm$ 2.4 & 0.66 $\pm$ 0.09 & 0.46 $\pm$ 0.11 \\
                    FGR \cite{zhou2016fast} & 22.2 $\pm$ 3.2 & 0.85 $\pm$ 0.06 & 0.24 $\pm$ 0.08 & 13.0 $\pm$ 2.6 & 0.57 $\pm$ 0.19 & 0.43 $\pm$ 0.20 & \cellcolor{GreenYellow!40}17.2 $\pm$ 2.3 & \cellcolor{GreenYellow!40}0.77 $\pm$ 0.08 & \cellcolor{GreenYellow!40}0.33 $\pm$ 0.11 \\
                    ICP \cite{rusinkiewicz2001efficient} & 22.7 $\pm$ 3.3 & 0.85 $\pm$ 0.06 & 0.24 $\pm$ 0.08 & 12.9 $\pm$ 2.6 & 0.57 $\pm$ 0.19 & 0.43 $\pm$ 0.20 & 16.8 $\pm$ 2.2 & 0.76 $\pm$ 0.09 & 0.35 $\pm$ 0.11 \\
                    Colored-ICP \cite{park2017colored} & \cellcolor{GreenYellow!40}{26.2 $\pm$ 3.1} & \cellcolor{Goldenrod!40}{0.89 $\pm$ 0.04} & \cellcolor{Goldenrod!40}{0.17 $\pm$ 0.06} & 13.4 $\pm$ 2.4 & \cellcolor{Goldenrod!40}{0.58 $\pm$ 0.19} & 0.40 $\pm$ 0.18 & 15.4 $\pm$ 2.1 & 0.71 $\pm$ 0.10 & 0.41 $\pm$ 0.13 \\
                    \algname-NR [{Ours}] & \cellcolor{Goldenrod!40}{26.3 $\pm$ 3.1} & \cellcolor{GreenYellow!40}{0.87 $\pm$ 0.05} & \cellcolor{Goldenrod!40}{0.17 $\pm$ 0.06} & \cellcolor{GreenYellow!40}15.4 $\pm$ 1.7 & 0.52 $\pm$ 0.12 & \cellcolor{GreenYellow!40}0.35 $\pm$ 0.05 & \cellcolor{WildStrawberry!40}24.8 $\pm$ 3.3 & \cellcolor{WildStrawberry!40}{0.83 $\pm$ 0.04} & \cellcolor{WildStrawberry!40}{0.22 $\pm$ 0.06} \\
                    \algname-R [{Ours}] & \cellcolor{WildStrawberry!40}{28.3 $\pm$ 2.9} & \cellcolor{WildStrawberry!40}{0.90 $\pm$ 0.04} & \cellcolor{WildStrawberry!40}{0.15 $\pm$ 0.06} & \cellcolor{Goldenrod!40}{16.4 $\pm$ 2.4} & \cellcolor{GreenYellow!40}0.57 $\pm$ 0.13 & \cellcolor{WildStrawberry!40}{0.31 $\pm$ 0.07} & \cellcolor{Goldenrod!40}{24.1 $\pm$ 3.1} & \cellcolor{Goldenrod!40}{0.82 $\pm$ 0.05} & \cellcolor{Goldenrod!40}{0.23 $\pm$ 0.06} \\
				\bottomrule
		\end{tabular}}
	\end{adjustbox}
\end{table*}

\subsection{Mobile-Robot Mapping}
\label{ssec:appendix_experiments_mobile_mapping}
\smallskip
\noindent\textbf{Photometric Evaluation.}
From \Cref{fig:photometric_performance_mobile_robot_rendered_images},
as highlighted by the green squares, \algname-R generates composite maps that are consistent with the ground-truth, unlike GaussReg and PhotoReg. The fused maps generated by GaussReg and PhotoReg contain conspicuous artifacts due to inaccurate registration of the individual maps created by the robots, especially in the \emph{Kitchen} scene. PhotoReg fails to sufficiently register the individual maps, resulting in blurry renderings, with few recognizable features, e.g., in the \emph{Workshop} scene. In the \emph{Apartment} scene, the rendered images from GaussReg contain duplicate objects, unlike those of \algname-R, which have accurate geometric detail.

\begin{table*}[th]
	\centering
	\caption{Photometric performance of GSplat registration algorithms for mobile-robot mapping.}
	\label{tab:baseline_photometric_performance_metrics_mobile_robot_mapping}
	\begin{adjustbox}{width=\linewidth}
		{\begin{tabular}{l | c c c | c c c | c c c }
				\toprule
                    & \multicolumn{3}{c |}{\emph{Kitchen}} & \multicolumn{3}{c |}{\emph{Workshop}} & \multicolumn{3}{c}{\emph{Apartment}} \\
				Methods & PSNR $\uparrow$ & SSIM  $\uparrow$ & LPIPS  $\downarrow$ & PSNR $\uparrow$ & SSIM  $\uparrow$ & LPIPS  $\downarrow$ & PSNR $\uparrow$ & SSIM  $\uparrow$ & LPIPS  $\downarrow$ \\
				\midrule
                    PhotoReg \cite{yuan2024photoreg} & 0.75 $\pm$ 0.05 & 11.6 $\pm$ 1.3 & 0.53 $\pm$ 0.06 & 0.78 $\pm$ 0.08 & 11.3 $\pm$ 1.3 &   0.48 $\pm$ 0.05 & 13.4 $\pm$ 1.5 & 0.61 $\pm$ 0.05 & 0.75 $\pm$ 0.04 \\
                    GaussReg \cite{chang2025gaussreg} & \cellcolor{GreenYellow!40}14.1 $\pm$ 1.9 & \cellcolor{GreenYellow!40}0.62 $\pm$ 0.06 & \cellcolor{GreenYellow!40}0.60 $\pm$ 0.11 & \cellcolor{GreenYellow!40}17.0 $\pm$ 2.8 & \cellcolor{Goldenrod!40}0.61  $\pm$ 0.05 & \cellcolor{GreenYellow!40}0.53 $\pm$ 0.11 & \cellcolor{GreenYellow!40}14.3 $\pm$ 1.6 & \cellcolor{GreenYellow!40}0.62 $\pm$ 0.05 & \cellcolor{GreenYellow!40}0.62 $\pm$ 0.04 \\
                    \algname-NR [\textbf{Ours}] & \cellcolor{WildStrawberry!40}19.3 $\pm$ 3.2 & \cellcolor{WildStrawberry!40}0.63 $\pm$ 0.05 & \cellcolor{WildStrawberry!40}0.40 $\pm$ 0.09 & \cellcolor{Goldenrod!40}19.9 $\pm$ 2.5 &  \cellcolor{GreenYellow!40}0.60 $\pm$ 0.04 & \cellcolor{Goldenrod!40}0.40 $\pm$ 0.08 & \cellcolor{Goldenrod!40}15.3 $\pm$ 1.5 &  \cellcolor{WildStrawberry!40}0.64 $\pm$ 0.04 & \cellcolor{WildStrawberry!40}0.55 $\pm$ 0.03   \\
                    \algname-R [\textbf{Ours}] & \cellcolor{Goldenrod!40}18.8 $\pm$ 2.8 & \cellcolor{Goldenrod!40}0.62 $\pm$ 0.05 & \cellcolor{Goldenrod!40}0.41 $\pm$ 0.08 & \cellcolor{WildStrawberry!40}20.3 $\pm$ 2.7 & \cellcolor{WildStrawberry!40}0.62 $\pm$ 0.04 & \cellcolor{WildStrawberry!40}0.38 $\pm$ 0.09 & \cellcolor{WildStrawberry!40}15.3 $\pm$ 1.4 & \cellcolor{Goldenrod!40}0.63 $\pm$ 0.04 & \cellcolor{WildStrawberry!40}0.55 $\pm$ 0.03  \\
				\bottomrule
		\end{tabular}}
	\end{adjustbox}
\end{table*}

\subsection{Ablations}
We examine the constituent registration steps in \algname, namely: the coarse Gaussian-to-Gaussian and fine photometric registration procedures, assessing the accuracy of the registration result generated by each procedure. We denote the variant of \algname with coarse registration performed without RANSAC and fine registration by \algname-CNR. Likewise, we denote the variant of \algname with coarse registration performed using RANSAC but without fine registration by \algname-CR. We compute the geometric and photometric performance metrics for each of these variants and report the results in \Cref{tab:baseline_geometric_performance_metrics_ablation} and \Cref{tab:baseline_photometric_performance_metrics_ablation}, respectively. We also report the performance of \algname-NR and \algname-R from \Cref{tab:baseline_geometric_performance_metrics} and \Cref{tab:baseline_photometric_performance_metrics} for easy reference.  From \Cref{tab:baseline_geometric_performance_metrics_ablation}, we note that the fine registration step in \algname notably improves the rotation error to sub-degree errors, achieving about $2$x smaller translation errors and in some cases, $100$x smaller translation errors. Likewise, the fine registration step generally results in much smaller scale errors, although not necessarily in all cases, as reflected in the \emph{Truck} scene.
Similarly, the variants of \algname with fine registration (i.e., \algname-NR and \algname-R) achieve notably higher photometric performance, especially in the \emph{Playroom} and \emph{Room} scenes, reported in \Cref{tab:baseline_photometric_performance_metrics_ablation}. 
In general, the coarse registration step brings corresponding objects in both GSplat maps into close proximity in the fused map. However, the resulting fused map lacks precise geometric detail, degrading its visual fidelity. After the coarse registration step, the fine registration procedure refines the transformation parameters for precise alignment of the individual maps, ultimately generating a photorealistic fused map.

Although \algname-CR does not always outperform RANSAC-GR in \Cref{tab:baseline_geometric_performance_metrics}, we observed empirically that the performance of RANSAC-GR has a high variance, posing a challenge for the fine registration step, which requires a sufficient number of corresponding features between rendered frames across the individual maps to compute a solution. Moreover, ICP and its variants tend to converge to a local optimum, close to the solution used for initialization. As a result, these methods generally fail to provide a sufficiently good initialization for the fine registration procedure. The coarse registration step in \algname relies significantly on the semantics extracted from the map to overcome these limitations, leveraging the inherent semantics to register corresponding objects at a sufficient level of accuracy for fine registration.

\begin{table*}[th]
	\centering
	\caption{Geometric Performance: Ablation of the Coarse Gaussian-to-Gaussian and Fine Photometric Registration in \algname.}
	\label{tab:baseline_geometric_performance_metrics_ablation}
	\begin{adjustbox}{width=\linewidth}
		{\begin{tabular}{l | c c c c | c c c c | c c c c}
				\toprule
                    & \multicolumn{4}{c |}{\emph{Playroom}} & \multicolumn{4}{c |}{\emph{Truck}} & \multicolumn{4}{c}{\emph{Room}} \\
				Methods & RE $\downarrow$ & TE  $\downarrow$ & SE $\downarrow$ & CT $\downarrow$ & RE $\downarrow$ & TE $\downarrow$ & SE $\downarrow$ & CT $\downarrow$ & RE $\downarrow$ & TE $\downarrow$ & SE $\downarrow$ & CT $\downarrow$ \\
				\midrule
                    \algname-CNR & 22.72 & {454.2} & {482.1} & \cellcolor{WildStrawberry!40}20.20 & {49.98} & {355.4} & 55.06 & \cellcolor{WildStrawberry!40}24.17 & {20.50} & {474.0} & {371.9} & \cellcolor{WildStrawberry!40}17.27 \\
                    \algname-CR & 21.15 & {324.2} & {51.94} & 20.47 & {0.804} & {7.691} & 7.744 & 26.32 & {24.07} & {381.8} & {155.1} & 17.58 \\
                    \algname-NR & \cellcolor{Goldenrod!40}0.348 & \cellcolor{Goldenrod!40}{4.860} & \cellcolor{Goldenrod!40}{0.282} & 41.16 & \cellcolor{Goldenrod!40}{0.511} & \cellcolor{Goldenrod!40}{8.07} & \cellcolor{Goldenrod!40}9.581 & 53.42 & \cellcolor{Goldenrod!40}{0.381} & \cellcolor{WildStrawberry!40}{2.648} & \cellcolor{WildStrawberry!40}{1.016} & 40.24 \\
                    \algname-R & \cellcolor{WildStrawberry!40}{0.170} & \cellcolor{WildStrawberry!40}{1.933} & \cellcolor{WildStrawberry!40}{0.170} & 39.73 & \cellcolor{WildStrawberry!40}{0.413} & \cellcolor{WildStrawberry!40}{6.845} & \cellcolor{WildStrawberry!40}{2.548} & 52.47 & \cellcolor{WildStrawberry!40}{0.237} & \cellcolor{Goldenrod!40}{3.289} & \cellcolor{Goldenrod!40}{2.673} & 39.71 \\
				\bottomrule
		\end{tabular}}
	\end{adjustbox}
\end{table*}

\begin{table*}[th]
	\centering
	\caption{Photometric Performance: Ablation of the Coarse Gaussian-to-Gaussian and Fine Photometric Registration in \algname.}
	\label{tab:baseline_photometric_performance_metrics_ablation}
	\begin{adjustbox}{width=\linewidth}
		{\begin{tabular}{l | c c c | c c c | c c c }
				\toprule
                    & \multicolumn{3}{c |}{\emph{Playroom}} & \multicolumn{3}{c |}{\emph{Truck}} & \multicolumn{3}{c}{\emph{Room}} \\
				Methods & PSNR $\uparrow$ & SSIM  $\uparrow$ & LPIPS  $\downarrow$ & PSNR $\uparrow$ & SSIM  $\uparrow$ & LPIPS  $\downarrow$ & PSNR $\uparrow$ & SSIM  $\uparrow$ & LPIPS  $\downarrow$ \\
				\midrule
                    \algname-CNR & {12.0 $\pm$ 3.2} & {0.67 $\pm$ 0.11} & {0.65 $\pm$ 0.17} & 11.4 $\pm$ 2.4 & 0.52 $\pm$ 0.11 & 0.63 $\pm$ 0.22 & 6.9 $\pm$ 5.9 & {0.72 $\pm$ 0.11} & {0.39 $\pm$ 0.15} \\
                    \algname-CR & {13.7 $\pm$ 3.4} & {0.65 $\pm$ 0.14} & {0.58 $\pm$ 0.19} & \cellcolor{Goldenrod!40}16.2 $\pm$ 2.2 & \cellcolor{Goldenrod!40}0.56 $\pm$ 0.14 & \cellcolor{WildStrawberry!40}0.31 $\pm$ 0.07 & 11.7 $\pm$ 1.9 & {0.48 $\pm$ 0.12} & {0.64 $\pm$ 0.10} \\
                    \algname-NR & \cellcolor{Goldenrod!40}{26.3 $\pm$ 3.1} & \cellcolor{Goldenrod!40}{0.87 $\pm$ 0.05} & \cellcolor{Goldenrod!40}{0.17 $\pm$ 0.06} & 15.4 $\pm$ 1.7 & 0.52 $\pm$ 0.12 & 0.35 $\pm$ 0.05 & \cellcolor{WildStrawberry!40}24.8 $\pm$ 3.3 & \cellcolor{WildStrawberry!40}{0.83 $\pm$ 0.04} & \cellcolor{WildStrawberry!40}{0.22 $\pm$ 0.06} \\
                    \algname-R & \cellcolor{WildStrawberry!40}{28.3 $\pm$ 2.9} & \cellcolor{WildStrawberry!40}{0.90 $\pm$ 0.04} & \cellcolor{WildStrawberry!40}{0.15 $\pm$ 0.06} & \cellcolor{WildStrawberry!40}{16.4 $\pm$ 2.4} & \cellcolor{WildStrawberry!40}0.57 $\pm$ 0.13 & \cellcolor{WildStrawberry!40}{0.31 $\pm$ 0.07} & \cellcolor{Goldenrod!40}{24.1 $\pm$ 3.1} & \cellcolor{Goldenrod!40}{0.82 $\pm$ 0.05} & \cellcolor{Goldenrod!40}{0.23 $\pm$ 0.06} \\
				\bottomrule
		\end{tabular}}
	\end{adjustbox}
\end{table*}

\subsection{Finetuning}
\label{ssec:finetuning}
\algname does not pre-process the local GSplat maps before registration of the maps, resulting in the retention of floaters in the fused map whenever floaters exist in the local maps. Here, we examine finetuning the fused map with rendered images from the local maps to remove visual artifacts, without requiring access to the data used in the training the local GSplat maps, i.e., we do not require access to the real-world camera images and poses. To finetune the fused map without access to the original dataset, we select camera poses expressed in the local frames of the local GSplat maps (e.g., randomly or via an informed approach) and render images from these maps at these camera poses. Subsequently, we transform the set of camera poses from their associated local frames to the frame of the fused map using the transformation parameters computed by \algname. We construct a finetuning dataset from the set of images and associated camera poses, which we use in finetuning the fused map.

\begin{figure*}[th]
    \centering
    \includegraphics[width=\linewidth]{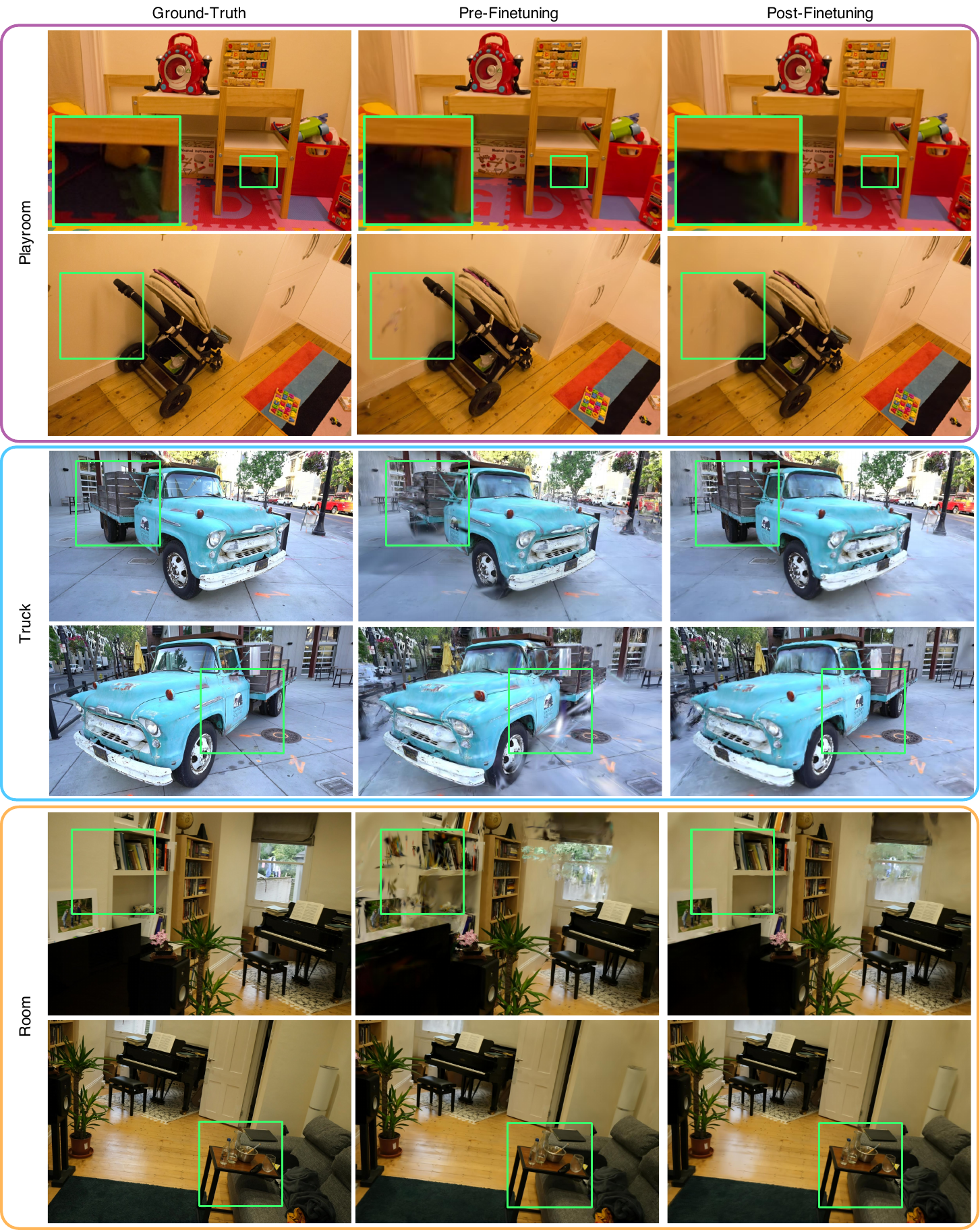}
    \caption{Rendered images from the fused GSplat maps generated by \algname-R before and after finetuning, in the \emph{Playroom}, \emph{Truck}, and \emph{Room} scenes. Finetuning improves the visual fidelity of the fused map, removing floaters and other artifacts.}
    \label{fig:photometric_performance_rendered_images_finetuning}
\end{figure*}

In \Cref{tab:finetuning_performance_metrics}, we provide the photometric scores of the fused GSplat map from \algname-R before and after finetuning and the ground-truth GSplat map. We train the ground-truth GSplat map using the combined training datasets used in training the local GSplat maps (i.e., the real-world camera images and poses, not the set of rendered images generated from the local GSplat maps), representing the ideal composite GSplat model. The computation time in \Cref{tab:finetuning_performance_metrics} represents the total training time for the ground-truth map and the total time used in finetuning the fused map. \Cref{tab:finetuning_performance_metrics} indicates that finetuning the fused map improves the PSNR, SSIM, and LPIPS scores compared to that of the pre-finetuned map. Specifically, in less than $90$ seconds, finetuning reduces the gap between the photometric scores of the ground-truth map and the photometric scores of the fused map by about $20\%$ to $40\%$. The relative improvements provided by finetuning the fused map depend on the finetuning data used, an area for future research. 
We provide rendered images from the fused GSplat map computed by \algname-R, before and after finetuning, and the corresponding images in the ground-truth fused map in \Cref{fig:photometric_performance_rendered_images_finetuning}. Across all three scenes, finetuning the fused map removes floaters and other artifacts, e.g., in the regions indicated by the green squares, ultimately resulting in higher PSNR and SSIM scores, as reported in \Cref{tab:finetuning_performance_metrics}.

\begin{table*}[th]
	\centering
	\caption{Photometric performance after finetuning \algname-R.}
	\label{tab:finetuning_performance_metrics}
	\begin{adjustbox}{width=\linewidth}
		{\begin{tabular}{l | c c c c | c c c c | c c c c }
				\toprule
                    & \multicolumn{4}{c |}{\emph{Playroom}} & \multicolumn{4}{c |}{\emph{Truck}} & \multicolumn{4}{c}{\emph{Room}} \\
				Methods & PSNR $\uparrow$ & SSIM  $\uparrow$ & LPIPS $\downarrow$ & CT $\downarrow$ & PSNR $\uparrow$ & SSIM  $\uparrow$ & LPIPS  $\downarrow$ & CT $\downarrow$ & PSNR $\uparrow$ & SSIM  $\uparrow$ & LPIPS  $\downarrow$ & CT $\downarrow$ \\
				\midrule
                    Ground-Truth & 36.3 $\pm$ 3.5 & 0.96 $\pm$ 0.03 & 0.09 $\pm$ 0.05 & 721.1 & 26.4 $\pm$ 1.4 & 0.89 $\pm$ 0.02 & 0.10 $\pm$ 0.01 & 601.7 & 34.1 $\pm$ 1.7 & 0.94 $\pm$ 0.02 & 0.12 $\pm$ 0.04 & 840.1 \\
                    Pre-Finetuning & 29.1 $\pm$ 3.3 & 0.91 $\pm$ 0.04 & 0.15 $\pm$ 0.06 & N/A & 16.8 $\pm$ 2.5 & 0.61 $\pm$ 0.10 & 0.30 $\pm$ 0.07 & N/A & 22.5 $\pm$ 2.5 & 0.79 $\pm$ 0.05 & 0.26 $\pm$ 0.06 & N/A \\
                    Post-Finetuning & 30.8 $\pm$ 2.6 & 0.92 $\pm$ 0.04 & 0.14 $\pm$ 0.06 & 72.69 & 21.1 $\pm$ 1.8 & 0.69 $\pm$ 0.1 & 0.23 $\pm$ 0.04 & 86.26 & 26.0 $\pm$ 3.6 & 0.83 $\pm$ 0.09 & 0.22 $\pm$ 0.08 & 79.78 \\
				\bottomrule
		\end{tabular}}
	\end{adjustbox}
\end{table*}

\end{appendices}

\end{document}